\documentclass[lettersize,journal]{IEEEtran}
\usepackage{amsmath,amsfonts}
\usepackage{algorithmic}
\usepackage{array}
\usepackage[caption=false,font=normalsize,labelfont=sf,textfont=sf]{subfig}
\usepackage{amsmath}
\usepackage{amssymb}
\usepackage{amsfonts}
\usepackage{booktabs, makecell, multirow, tabularx}
\usepackage{textcomp}
\usepackage{stfloats}
\usepackage{url}
\usepackage{verbatim}
\usepackage{graphicx}
\usepackage{xcolor}
\usepackage{bbm}
\usepackage{authblk}

\usepackage{lipsum}
\newcommand\mytab[1]{\begin{tabular}[t]{@{}c@{}} #1 \end{tabular}}
\newcommand\mc[2]{\multicolumn{#1}{c}{#2}}

\newcommand{\nk}{\kern-0.1em}

\def\BibTeX{{\rm B\kern-.05em{\sc i\kern-.025em b}\kern-.08em
    T\kern-.1667em\lower.7ex\hbox{E}\kern-.125emX}}
\usepackage{balance}
\begin{document}
\title{
DD-rPPGNet: De-interfering and Descriptive Feature Learning for Unsupervised rPPG Estimation  
}

\author[a,b]{Pei-Kai Huang}
\author[b]{Tzu-Hsien Chen}
\author[b]{Ya-Ting Chan}
\author[b]{Kuan-Wen Chen}
\author[b]{Chiou-Ting Hsu}
\affil[a]{College of Computer and Cyber Security, Fujian Normal University, Fuzhou, China}
\affil[b]{Department of Computer Science, National Tsing Hua University, Taiwan

\thanks{Manuscript received August 15, 2024; revised February 17, 2025; accepted April 20, 2025. The associate editor coordinating the review of this manuscript and approving it for publication was Prof. Zhen Lei.
\textit{(Corresponding author: Chiou-Ting Hsu.)}}

\thanks{The authors are with  National Tsing Hua University.}

\thanks{The supplementary material is available at https://github.com/Pei-KaiHuang/TIFS2025-DD-rPPGNet/blob/main/Supplementary-Material-of-DD-rPPGNet-TIFS-2025.pdf, provided by the author.}

 
}

\markboth{Journal of IEEE Transactions on Information Forensics and Security, April~2025}%
{How to Use the IEEEtran \LaTeX \ Templates}

\maketitle

\begin{abstract}

Remote Photoplethysmography (rPPG) aims to measure physiological signals and Heart Rate (HR) from facial videos.
Recent unsupervised rPPG estimation methods have shown promising potential in estimating rPPG signals from facial regions without relying on ground truth rPPG signals.
However, these methods seem oblivious to interference existing in rPPG signals and still result in unsatisfactory performance. In this paper, we propose a novel De-interfered and Descriptive rPPG Estimation Network (DD-rPPGNet) to eliminate the interference within rPPG features for learning genuine rPPG signals.
First, we investigate the characteristics of local spatial-temporal similarities of interference and design a novel unsupervised model to estimate the interference.
Next, we propose an unsupervised de-interfered method to learn genuine rPPG signals with two stages. In the first stage, we estimate the initial rPPG signals by contrastive learning from both the training data and their augmented counterparts. 
In the second stage, we use the estimated interference features to derive de-interfered rPPG features and encourage the rPPG signals to be distinct from the interference. 
In addition, we propose an effective descriptive rPPG feature learning by developing a strong 3D Learnable Descriptive Convolution (3DLDC) to capture the subtle chrominance changes for enhancing rPPG estimation. 
Extensive experiments conducted on five rPPG benchmark datasets demonstrate that the proposed DD-rPPGNet outperforms previous unsupervised rPPG estimation methods and achieves competitive performances with state-of-the-art supervised rPPG methods.
The code is available at: https://github.com/Pei-KaiHuang/TIFS2025-DD-rPPGNet
   
\end{abstract}


\section{Introduction}
\label{sec:intro}

\begin{figure} [t]
   \begin{minipage}[b]{1\linewidth}
    \centering
    \begin{tabular}{c c}
        \begin{minipage}{.9\textwidth}
            \includegraphics[width=\linewidth]{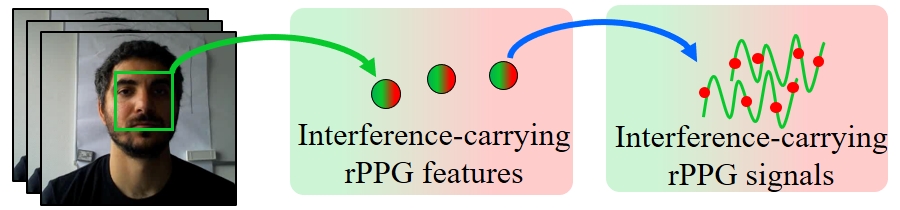}
        \end{minipage} \\ (a)
        \\        
        \begin{minipage}{.9\textwidth}
            \includegraphics[width=\linewidth]{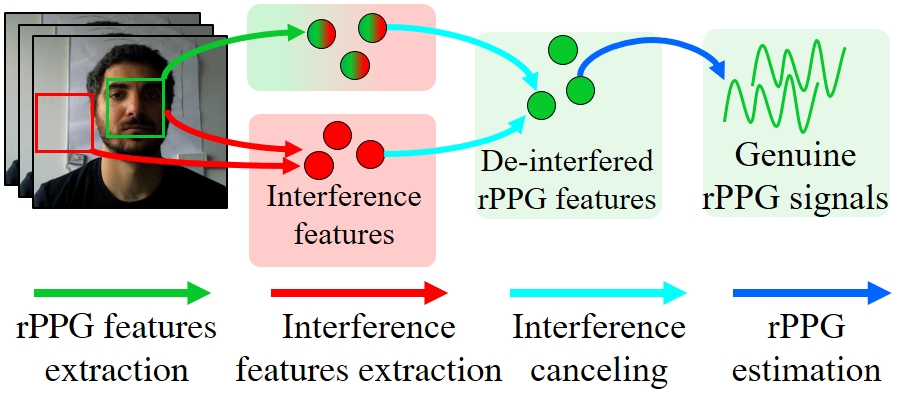}
        \end{minipage} \\ (b)
        \\ 
    \end{tabular}   
  \end{minipage} 
\caption{    
Illustration of de-interferenced feature learning in unsupervised rPPG estimation. 
(a) Previous methods often disregard the existence of interference in rPPG features and tend to extract interference-carrying rPPG signals.
(b) In this paper, we propose to model the interference features 
and use the estimated interference features to derive de-interfered rPPG features for learning genuine rPPG signals.
}
\label{fig:idea}   
\end{figure}

\IEEEPARstart{R}{emote} photoplethysmography (rPPG) aims to measure physiological signals without any skin contact \cite{chen2018deepphys, Spetlik2018VisualHR} by analyzing the facial chrominance changes reflected on skin \cite{yu2019remote, niu2019rhythmnet, yu2019remote_Phys}, and is used to capture  the heart rate related information \cite{lee2020meta, liu2020multi, niu2020video, nowara2020near, tsou2020multi}.
Many deep learning-based methods \cite{ nowara2021benefit, lu2021dual, yu2021transrppg, huang2021spatio, yu2022physformer, hsieh2022augmentation, chung2022domain}, through the adoption of supervised learning techniques, have achieved significant breakthroughs in rPPG estimation. 
These supervised rPPG estimation methods rely on paired facial videos and ground truth rPPG signals to learn the estimation of rPPG signals from facial videos. However, collection of paired facial videos and ground truth rPPG signals is very costly, and also their quality highly depends on the adopted contact sensors, the involved subjects, and the environmental settings. 
Therefore, existing datasets, in terms of their quantity and quality, remain insufficient for developing robust supervised rPPG estimation methods.

In contrast, unsupervised rPPG estimation methods \cite{gideon2021way,sun2022contrast,speth2023non,yue2023facial} have been developed to estimate rPPG signals directly from face videos without referring to paired ground truth rPPG signals.
These methods are developed based on several assumptions about the nature of rPPG signals.
For example, in \cite{gideon2021way}, the authors adopted contrastive learning, under the observation of temporal similarity of rPPG signals and the typical range of  human heart rate (within  40 to 250 beats per minute (bpm)), to develop a unsupervised rPPG estimation method.
Next, the authors in \cite{sun2022contrast} further included the observation of spatial similarity within each individual video and the observation of dissimilarity across different videos to refine the contrastive learning in unsupervised rPPG estimation. 
Furthermore,  in \cite{yue2023facial}, the authors proposed to augment the training data by using dissimilar signal frequencies for enhancing unsupervised rPPG estimation.
In addition, the authors in \cite{speth2023non} proposed a non-contrastive unsupervised rPPG estimation approach to learn concentrated energy between the heart rate ranges.
Although these unsupervised rPPG estimation methods may perform well on datasets with minor interference, they usually fail to estimate accurate rPPG signals in datasets containing challenging interference.
Since facial videos are inevitably corrupted by various interference, such as head motions \cite{chen2018deepphys,maity2022robustppg}, facial expressions \cite{zhang2016multimodal}, video compression artifacts \cite{yu2019remote}, and lighting variations \cite{magdalena2018sparseppg}, rPPG estimation is very sensitive to these interferencing sources. As shown in Figure \ref{fig:idea} (a), previous unsupervised rPPG estimation methods \cite{gideon2021way,sun2022contrast,speth2023non,yue2023facial} often disregard the existence of interference in rPPG features and tend to extract interference-carrying rPPG signals. 
In Section \ref{sec:Involving Interferences}, we will show that the rPPG signals estimated from existing unsupervised rPPG estimation methods, such as \cite{sun2022contrast}, exhibit a notably positive correlation with the interference in non-facial background regions.
To eliminate interference in rPPG features, previous supervised methods \cite{lu2021dual, nowara2021benefit, kang2022transppg, maity2022robustppg} mostly relied on ground truth rPPG signals to model interference for learning genuine rPPG signals.
As to unsupervised rPPG estimation methods, which involve no paired ground truth rPPG signals, the task of extracting de-interfered rPPG signals is much more challenging compared to supervised methods. 
In addition, estimation of rPPG signals is a nontrivial task because they are estimated from subtle chrominance changes reflected on skin. 
To capture these subtle skin color changes, previous supervised methods \cite{yu2020autohr,yu2022physformer,zhao2021video} included several pre-defined local descriptors into vanilla 3DCNN, such as 3D Temporal Difference Convolution (TDC) \cite{yu2020autohr,yu2022physformer}, and 3D Central Difference Convolution (3DCDC) \cite{zhao2021video}, to learn descriptive features for enhancing rPPG estimation.
Notably, inclusion of learning descriptive rPPG features in unsupervised rPPG estimation remains unexplored.
Therefore, when dealing with facial videos corrupted by various interferences, rPPG estimation is indeed doubly challenging to unsupervised methods.  

In this paper, to address the above-mentioned challenges in unsupervised rPPG estimation, we propose a novel De-interfered and Descriptive rPPG Estimation Network (DD-rPPGNet) to eliminate interference from interference-carrying rPPG features for learning genuine rPPG signals. 
Figure \ref{fig:idea} (b) illustrates our main idea. 
As noted in \cite{nowara2021benefit}, interference in foreground and background are approximately similar if they are resulted from the same source. In Section  \ref{sec:Distinguishing Characteristics of Noises}, we will also verify that foreground and background interference signals indeed exhibit local spatial-temporal similarities.
Therefore, we adopt these characteristics of interference between foreground facial region and background non-facial region to develop an unsupervised interference estimation method.  
Next, we propose an unsupervised de-interfered rPPG estimation to learn genuine rPPG signals with two stages. 
In the first stage, we use contrastive learning to estimate the initial rPPG signals from both the training data and their weakly augmented counterparts. 
In the second stage, we use the estimated interference features to derive de-interfered rPPG features and also encourage the refined rPPG signals to be distinct from the interference.
In addition, to tackle the challenge caused by subtle chrominance changes, we further propose an effective descriptive rPPG feature learning by extending our previously developed 2DLDC \cite{huang2022learnable} to develop a 3D Learnable Descriptive Convolution (3DLDC) to capture the subtle temporal and chrominance changes for enhancing rPPG estimation. 
We conduct extensive experiments on five public rPPG databases to evaluate the effectiveness of the proposed DD-rPPGNet.
Our experimental results on intra-domain and cross-domain testing demonstrate that the proposed DD-rPPGNet not only outperforms previous unsupervised rPPG estimation methods  but also achieves competitive performance to state-of-the-art supervised rPPG estimation methods.

Our contributions are summarized as follows:

\noindent 
$\bullet$  
We propose a novel model DD-rPPGNet, focusing on describing interference characteristics and on eliminating interference from rPPG features, to learn genuine rPPG signals for unsupervised rPPG estimation. 

\noindent 
$\bullet$ 
Based on our analysis of local spatial-temporal interference similarities between foreground and background regions, we propose a novel unsupervised method for interference estimation to substantially improve rPPG estimation. 

\noindent 
$\bullet$ 
To effectively capture subtle chrominance changes on facial skin, we additionally introduce a robust 3D Learnable Descriptive Convolution (3DLDC) by incorporating learnable local descriptors into 3DCNN for learning descriptive rPPG features.  
 
\noindent 
$\bullet$
Extensive experiments demonstrate that DD-rPPGNet outperforms previous unsupervised rPPG estimation methods and achieves competitive performance with state-of-the-art supervised rPPG estimation methods.

\section{Related Work }  

\subsection{Supervised rPPG Estimation} 
Supervised rPPG estimation focuses on learning the rPPG estimation model from paired facial videos and ground truth rPPG signals. 
Earlier approaches have developed various end-to-end architectures to directly learn rPPG relevant features.
For example, in \cite{chen2018deepphys}, the authors proposed the first end-to-end system to capture motion information between consecutive frames for robust HR measurement under significant moving scenarios. 
Some other end-to-end methods designed different networks, such as deep spatio-temporal network \cite{yu2019remote_Phys,yu2019remote}, siamese-rPPG network \cite{tsou2020siamese}, and recent  PhysFormer \cite{yu2022physformer}  to improve the accuracy of rPPG estimation.
In \cite{tsou2020multi,hsieh2022augmentation}, the authors proposed to generate augmented training data through extracting and swapping rPPG signals across different videos.

Since remote estimation of physiological signals on skins is highly sensitive to various interference, such as head motions \cite{chen2018deepphys,maity2022robustppg}, facial expressions \cite{zhang2016multimodal}, video compression artifacts \cite{yu2019remote}, and lighting variations \cite{magdalena2018sparseppg},  
many methods \cite{song2021pulsegan,lu2021dual,du2023dual,nowara2021benefit,maity2022robustppg} relied on ground truth rPPG signals to model interference for learning de-interfered rPPG signals.
In particular, in \cite{song2021pulsegan,lu2021dual,du2023dual}, the authors proposed to adopt generative adversarial learning to model the noise distribution of interference.
Also, in \cite{nowara2021benefit}, the authors proposed to learn interference co-existing in facial and non-facial regions to suppress their shared covariates and to amplify the rPPG signal information.
In \cite{maity2022robustppg}, the authors proposed to generate the motion distortion for filtering out the motion-induced measurements.

Moreover, to capture subtle chrominance changes on facial skin, many methods \cite{yu2020autohr,yu2022physformer,zhao2021video} included pre-defined and fixed local descriptors into vanilla 3DCNN, such as 3D Temporal Difference Convolution (TDC) \cite{yu2020autohr,yu2022physformer} and 3D Central Difference Convolution (3DCDC) \cite{zhao2021video}, to enhance the representation capacity of vanilla 3DCNN for learning descriptive rPPG features.
In addition, as noted in \cite{huang2024survey,huang2025slip}, rPPG estimation can also assist in detecting live faces and distinguishing them from facial spoof attacks, as 3D mask facial attacks lack intrinsic rPPG information \cite{huang2024one,huang2025channel,huang2023towards}.

\subsection{Unsupervised rPPG Estimation} 

Unlike supervised method, unsupervised rPPG estimation aims to learn rPPG estimators solely from facial videos without relying on ground truth rPPG signals.
In \cite{gideon2021way}, building upon on the observation that rPPG signals exhibit slight variations over small time intervals and that human heart rates generally fall between 40 and 250 beats per minute (bpm), the authors proposed 
an unsupervised rPPG estimation via contrastive learning between original videos and their resampled counterparts.
Next, in \cite{sun2022contrast}, by further including the observations that rPPG signals exhibit small variations across different spatial regions of the same video and exhibit larger variations across different videos, the authors extended the findings in \cite{gideon2021way} and proposed adopting spatio-temporal constraints to develop an unsupervised method.
Furthermore, in \cite{yue2023facial}, the authors proposed learning the rPPG signals from augmented videos with dissimilar signal frequencies.
In \cite{speth2023non}, the authors proposed the first non-contrastive learning framework to learn concentrated energy within the heart rate range for unsupervised rPPG estimation.
In addition, \cite{huang2024fully} proposed utilizing test-time adaptation techniques \cite{huang2023test} for rPPG estimation to enhance the model’s ability to adapt to target data with new domain information and unknown heart rates.
While these unsupervised rPPG estimation methods \cite{gideon2021way,sun2022contrast,speth2023non,yue2023facial} have shown promising potential, inclusion of de-interfering and descriptive feature featuring remains unexplored.

\section{Background and Motivation} 
\label{sec: background_observation}  
  
\subsection{ Interference Model for rPPG Estimation }  
\label{sec:Interference model}

In \cite{nowara2021benefit}, the authors explicitly modeled the observed intensities in foreground facial component $x^{fg}(p)$ and in background (i.e., non-facial component) $x^{bg}(p)$ of each pixel $p$ by,
\begin{normalsize}  
\begin{eqnarray}
\label{eq:fg_intensity}    
x^{fg}(p) &=& base(p) + {n}^{fg } +  {r},
\\
\label{eq:fg_intensity2}   
x^{bg}(p) &=& base(p) + {n}^{bg},
\end{eqnarray}
\end{normalsize}
 
\noindent 

where $base(p)$ denotes the base intensity of the pixel $p$, 
$r$ denotes the rPPG signal, and
${n}^{fg}$ and $ {n}^{bg}$ denote the interference in the facial and non-facial regions, respectively.
Note that, as mentioned in \cite{nowara2021benefit,kang2022transppg}, the interference ${n}^{fg}$ and $ {n}^{bg}$ in Equations \eqref{eq:fg_intensity} and \eqref{eq:fg_intensity2} do not cover random noises (e.g., camera sensor noises), because these noises are randomly and independently distributed and thus are not included in the model.

From Equations \eqref{eq:fg_intensity} and \eqref{eq:fg_intensity2}, if the interference signals ${n}^{fg}$ and $ {n}^{bg}$ are resulted from the same source, e.g., video compression or illumination variations from a flickering light bulb, then we can approximate the foreground interference by the the background interference, i.e., ${n}^{fg} \approx {n}^{bg} $. 
Therefore, in \cite{nowara2021benefit,kang2022transppg}, the authors proposed to learn the interference ${n}^{bg}$ from non-facial regions $x^{bg}$ and then use ${n}^{bg}$ to cancel the interference in $x^{fg}$ for enhancing learning the rPPG signals $r$.
Similarly, other supervised methods either utilized generative adversarial learning for generating interference \cite{song2021pulsegan,lu2021dual,du2023dual} or adopted motion distortion model \cite{maity2022robustppg}  
to eliminate the interference for enhancing the rPPG estimation.

Unlike the supervised methods \cite{liu2020multi, song2021pulsegan, nowara2021benefit, lu2021dual, maity2022robustppg, du2023dual}, which rely on the ground truth rPPG signals to learn the de-interfered rPPG signals, existing unsupervised methods \cite{gideon2021way,sun2022contrast,speth2023non,yue2023facial} 
are mostly built upon contrastive learning based on numerous assumptions, such as rPPG spatial-temporal similarity and HR range constraint, to directly estimate rPPG signals. 
Since these unsupervised methods \cite{gideon2021way,sun2022contrast,speth2023non,yue2023facial} are
oblivious to the rPPG interference model, their estimated rPPG signals $\hat{r}$ are highly susceptible to the interference ${n}^{fg}$, i.e., $\hat{r} \cong {r} + {n}^{fg} $. 
In Section \ref{sec:Involving Interferences}, we will justify this claim by showing that 
the rPPG signals estimated from existing unsupervised rPPG estimation methods are indeed positively and significantly correlated with the interference ${n}^{bg}$ in the non-facial background region. In Sections \ref{sec:Distinguishing Characteristics of Noises} and \ref{sec:Characteristics of rPPG}, we will then discuss the characteristics of ${n}^{fg} $ and ${n}^{fg}$ and those of rPPG signals $r$, respectively, and will then develop a novel method in Section \ref{sec:Proposed_method} to improve estimating the genuine and de-interfered rPPG signals $r$ under unsupervised learning scenario.

\begin{figure}[t]  {
    \centering
    \includegraphics[width=7.5cm]{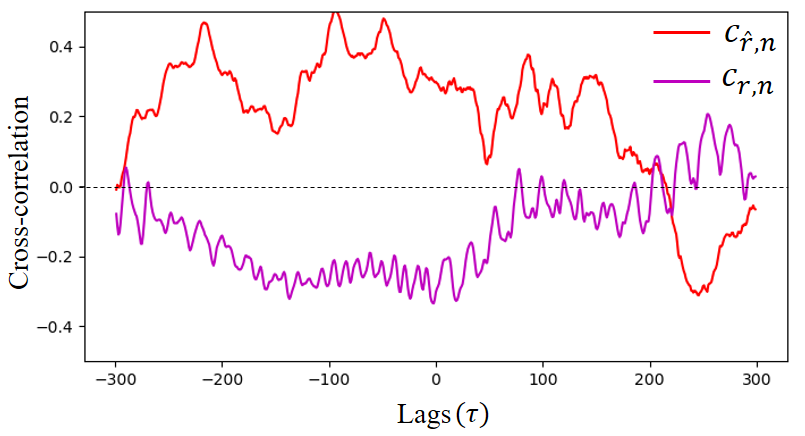} 
\caption{The two curves ${c}_{\hat{r}, {n}}[\tau]$ (in red color) and ${c}_{{r}, {n}}[\tau]$ (in purple color) correspond to the running correlations between the estimated rPPG signal $\hat{r}$ and the interference ${n}^{bg}$ and between the ground truth rPPG signal ${r}$ and the interference ${n}^{bg}$, respectively. 
This experiment shows that, $\hat{r}$ and ${n}^{bg}$ have significantly positive correlations ${c}_{\hat{r}, {n}}[\tau]$, whereas ${r}$ and ${n}^{bg}$ have negative correlations ${c}_{{r}, {n}}[\tau]$.
\label{fig:signal_cc_noisy_rPPG_and_noise}}}
\end{figure}

\subsection{rPPG Estimation Involving Interference}
\label{sec:Involving Interferences}

As mentioned in \cite{nowara2021benefit}, since foreground and background interference signals usually come from the same source, we can readily use ${n}^{bg}$ to approximate ${n}^{fg}$ by ${n}^{fg} \approx {n}^{bg}$.
Here, we will first analyze the correlation between the estimated rPPG signals $\hat{r}$ and the interference signals ${n}^{bg}$ and then will investigate whether the rPPG signals $\hat{r}$ are corrupted by interference or not.

\begin{figure}[t]  
    \centering
    \begin{tabular}{c} 
     {\includegraphics[width=7.5cm]{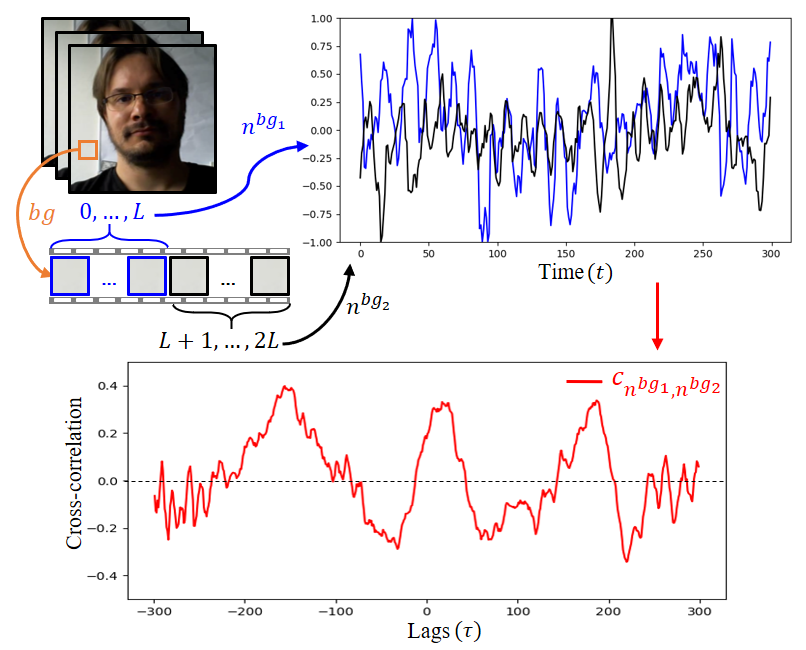}}   
    \end{tabular} 
\caption{  
The two interference signals ${n}^{bg_{1}}$ and ${n}^{bg_{2}}$, extracted from two non-overlapped and equal-length clips of the same non-facial region $bg$, not only have similar waveforms but also have strong positive correlation ${c}_{{n^{bg_1}},{n^{bg_2}}}[\tau]$.
} 

\label{fig:temporal similarity} 
\end{figure}  

First, we adopt the pre-trained model \cite{sun2022contrast} as the rPPG estimator to extract both the rPPG signals $\hat{r}$ and the background interference ${n}^{bg}$
from foreground facial region and background non-facial region, respectively.
Next, we measure the running correlation ${c}_{\hat{r}, {n}}$ between $\hat{r}$ and ${n}^{bg}$ by,
\begin{normalsize}  
\begin{eqnarray}
\label{eq:running correlation signal}   
 {c}_{\hat{r}, {n}}[\tau] &=& NC (\hat{r}[v-\tau],  {n}^{bg}[v]), \nonumber \\
 &=& NC (\hat{r}[\bar{v}],  {n}^{bg}[v]),  
\end{eqnarray}
\end{normalsize} 
  
\noindent 
where $NC(a[v],b[v]) = \frac{\sum_l^L a[l] \cdot b[l]}{\sqrt{\sum_l^L (a[l])^2  }\sqrt{\sum_l^L (b[l])^2  }}$ 
is the normalized correlation between the two signals $a[v]$ and $b[v]$ of length $L$, $\hat{r}[\bar{v}] = \hat{r}[v-\tau]$ is the shifted signal of $\hat{r}[v]$ by $\tau$ samples, and $\tau \in \{-(L-1),\cdots,0,\cdots,L-1\}$ indicates the running lag.

Assuming the estimated rPPG signals $\hat{r}$ are corrupted by the interference ${n}^{fg}$, i.e., $\hat{r} \cong {r} + {n}^{fg}$,  then we rewrite Equation \eqref{eq:running correlation signal} by substituting $\hat{r}$ with ${r} + {n}^{fg} $ and have:  

\begin{normalsize}  
\begin{eqnarray}
\begin{split}
\label{eq:rewrite running correlation}   
 {c}_{\hat{r},{n}}[\tau]  
 & \cong NC ( {r}[v-\tau] +  {n}^{fg}[v-\tau] ,  {n}^{bg}[v])   \\
 & \cong NC ( {r}[\bar{v}] +  {n}^{fg}[\bar{v}] ,  {n}^{bg}[v])   \\
 & \cong \alpha  NC ({r}[\bar{v}],  {n}^{bg}[v]) +  \beta  NC ({n}^{fg}[\bar{v}],  {n}^{bg}[v]),
\end{split}
\end{eqnarray}
\end{normalsize} 
 
\noindent

where $\alpha = \frac{\sqrt{\sum_l^{L} ({r}[l])^2}}
{\sqrt{\sum_l^{L} ({r}[l]+ {n}^{fg}[l])^2}}$,  $\beta = \frac{\sqrt{\sum_l^{L} ({n}^{fg}[l])^2}}
{\sqrt{\sum_l^{L} ({r}[l]+{n}^{fg}[l])^2}}$, 
and ${r}[l]$ and ${n}^{fg}[l]$ are the $l$-th samples of ${r}[\bar{v}]$ and ${n}^{fg}[\bar{v}]$, respectively.
Please see the supplementary material for detailed derivations of Equation \eqref{eq:rewrite running correlation}. 

In Equation \eqref{eq:rewrite running correlation}, the correlation between $\hat{r}$ and ${n}^{bg}$ can be decomposed into  two terms. 
The first one is the correlation between the genuine rPPG signal ${r}$ and the estimated interference ${n}^{bg}$; and the second one is the correlation between the two interference signals ${n}^{fg}$ and ${n}^{bg}$.
Therefore, if an estimated rPPG signal $\hat{r}$ involves no or little interference, i.e., $\hat{r} \cong {r} $, then $\hat{r}$ should be independent of the interference ${n}^{bg}$ (the first term in Equation \eqref{eq:rewrite running correlation}) and should have little correlation between  ${n}^{fg}$ and ${n}^{bg}$ (the second term in Equation \eqref{eq:rewrite running correlation}).
In Figure \ref{fig:signal_cc_noisy_rPPG_and_noise}, we show the running correlation   ${c}_{\hat{r}, {n}}[\tau]$ between $\hat{r}$ and ${n}^{bg}$ and also the running correlation ${c}_{{r}, {n}}[\tau]$ between the ground truth rPPG signal $r$ and ${n}^{bg}$ measured by Equation \eqref{eq:running correlation signal}. 
From Figure \ref{fig:signal_cc_noisy_rPPG_and_noise}, we see that ${r}$ and ${n}^{bg}$ are indeed negatively correlated because of their independent nature. 
However, the positive correlations ${c}_{\hat{r}, {n}}[\tau]$ between $\hat{r}$ and ${n}^{bg}$ show that the estimated rPPG signals $\hat{r}$ are highly corrupted by the interference ${n}^{fg}$.

\begin{figure} 
    \centering
    \begin{tabular}{c c}
        \begin{minipage}{.43\textwidth}
            \includegraphics[width=\linewidth]{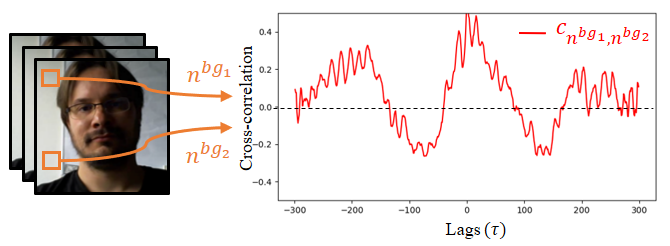}
        \end{minipage} \\ (a)
        \\        
        \begin{minipage}{.43\textwidth}
            \includegraphics[width=\linewidth]{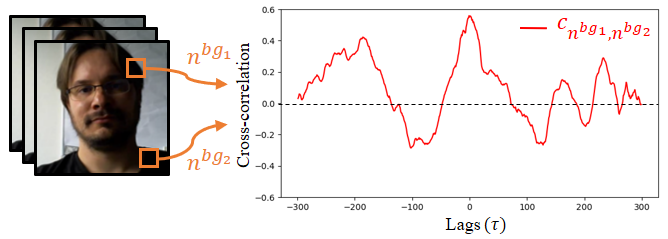}
        \end{minipage} \\ (b)
        \\
        \begin{minipage}{.43\textwidth}
            \includegraphics[width=\linewidth]{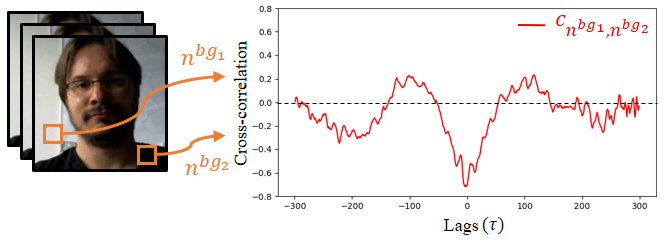}
        \end{minipage} \\ (c)
        \\
    \end{tabular}
\caption{  
In (a) and (b), the interference signals ${n}^{bg_{1}}$ and ${n}^{bg_{2}}$, extracted from (a) two well-illuminated regions and (b) two dimly-illuminated regions, have a high positive correlation ${c}_{{n^{bg_1}},{n^{bg_2}}}[\tau]$. In contrast, in (c), the signals ${n}^{bg_{1}}$ and ${n}^{bg_{2}}$, extracted from two differently illuminated regions, have a negative correlation ${c}_{{n^{bg_1}},{n^{bg_2}}}[\tau]$.
} 
\label{fig:local similarity} 
\end{figure}

\subsection{Characteristics of Interference Signals}
\label{sec:Distinguishing Characteristics of Noises}

\subsubsection{Local Spatial-Temporal Similarities of Interference}

As observed in \cite{nowara2021benefit}, each facial video often has interference originating from the same source, such as compression artifacts \cite{yu2019remote} or lighting variations \cite{magdalena2018sparseppg}.
Because these sources of interference are relatively stable across the temporal domain, it is reasonable to assume that the interference exhibits temporal similarity within a single video.
On the other hand, when videos are captured under natural light conditions with uneven illuminations, the resultant interference on different spatial regions (e.g., well-illuminated and dimly illuminated regions) usually exhibits varying characteristics.
Since the interference signals in different spatial regions originate from the same source, they should still exhibit local spatial similarities.
Therefore, from the above two observations, we infer that the interference signals in rPPG exhibit local spatial-temporal similarities and will experimentally verify these characteristics below.

\subsubsection{Experimental Verification}
 
To evaluate the local spatial and temporal characteristics of interference, we adopt the off-the-shelf rPPG estimator \cite{sun2022contrast} to extract signals from different regions within a single video and across multiple rPPG datasets.

\paragraph{Within a Single Video} 
In Figures \ref{fig:temporal similarity} and \ref{fig:local similarity}, we extract interference signals ${n}^{bg_1}$ and ${n}^{bg_2}$ of non-facial regions $bg_1$ and $bg_2$ in a sample video from \textbf{COHFACE} dataset \cite{heusch2017reproducible} and generate the running correlation ${c}_{{n^{bg_1}},{n^{bg_2}}}[\tau]$ to analyze the relationship between ${n}^{bg_1}$ and ${n}^{bg_2}$ during the time lag $\tau$. 
Note that, because we extract the signals from non-facial regions $x^{bg}$, these signals capture no rPPG information but reflect only the interference ${n}^{bg}$.

In Figure \ref{fig:temporal similarity}, we first show that the interference signals ${n}^{bg_1}$ and ${n}^{bg_2}$, extracted from two non-overlapped clips of the same non-facial region, display similar waveform and have a strong positive correlation across temporal domain. 
Next, in Figure \ref{fig:local similarity} (a)-(c), we measure the running correlation between the signals ${n}^{bg_1}$ and ${n}^{bg_2}$ extracted from two well-illuminated regions, two dimly-illuminated regions, and two differently illuminated regions, respectively. The results in Figure \ref{fig:local similarity} (a) and (b) show that the extracted interference signals are also highly similar and have a high positive correlation. 
By contrast, in Figure \ref{fig:local similarity} (c), the two interference signals, extracted from the same video under different lighting conditions, exhibit a significantly negative correlation. 
Consequently, the results in Figures \ref{fig:temporal similarity} and \ref{fig:local similarity} experimentally demonstrate that the interference signals in rPPG model
exhibit local spatial-temporal similarity.

\begin{figure} [t] 
    \begin{minipage}[b]{1.0\linewidth}
      \centering{ 
      \includegraphics[width=8.cm]{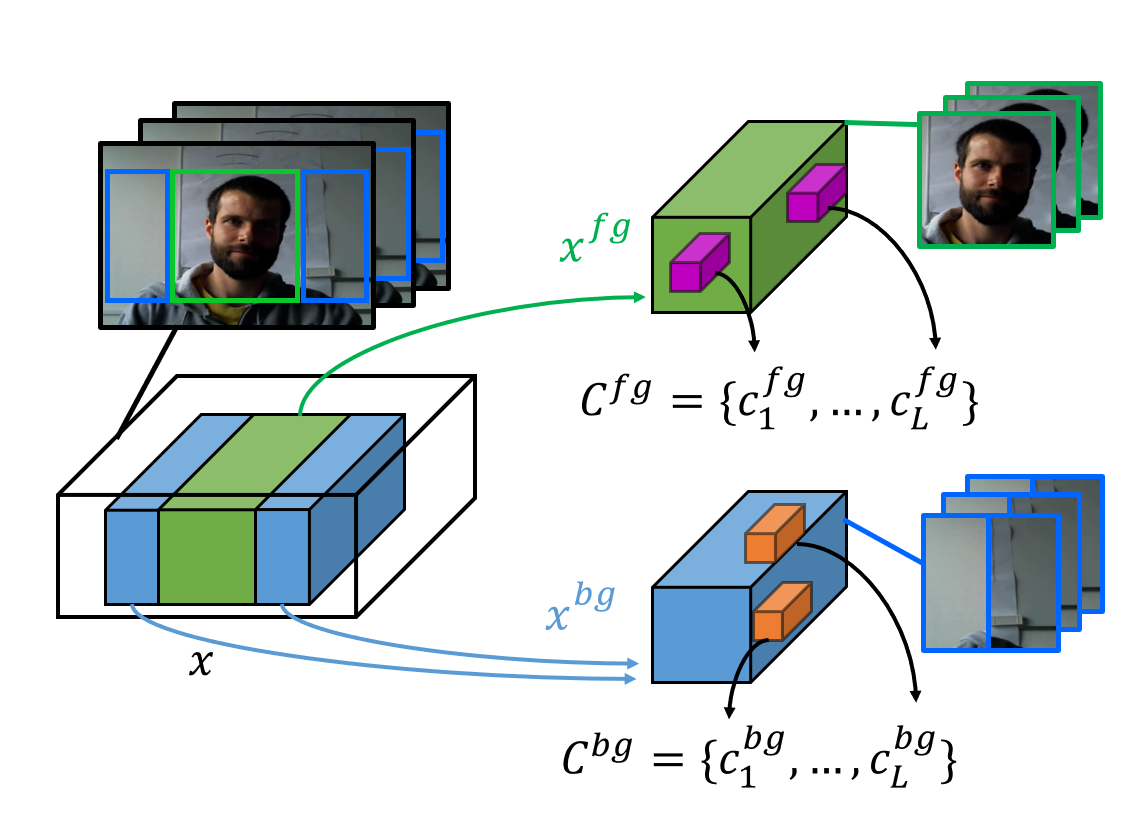}
      }
    \end{minipage} 
\caption{   
Illustration of sampling the set of foreground clips $C^{fg}$ and the set of background clips $C^{bg}$ from an input video $x$. We first locate the foreground facial region $x^{fg}$ and the background non-facial region $x^{bg}$ in $x$. Next, we randomly sample $L$ clips from different foreground facial regions and from different background non-facial regions to generate $C^{fg}$ and $C^{bg}$, respectively. All the foreground clip $c^{fg} \in C^{fg}$ and the background clip $c^{bg} \in C^{bg}$ have $\Delta_t$ frames and are with fixed size of $h \times w$.}

 \label{fig:preprocessing}  
\end{figure}

\begin{figure} 
    \centering
    \begin{tabular}{c c}
        \begin{minipage}{.48\textwidth}
            \includegraphics[width=\linewidth]{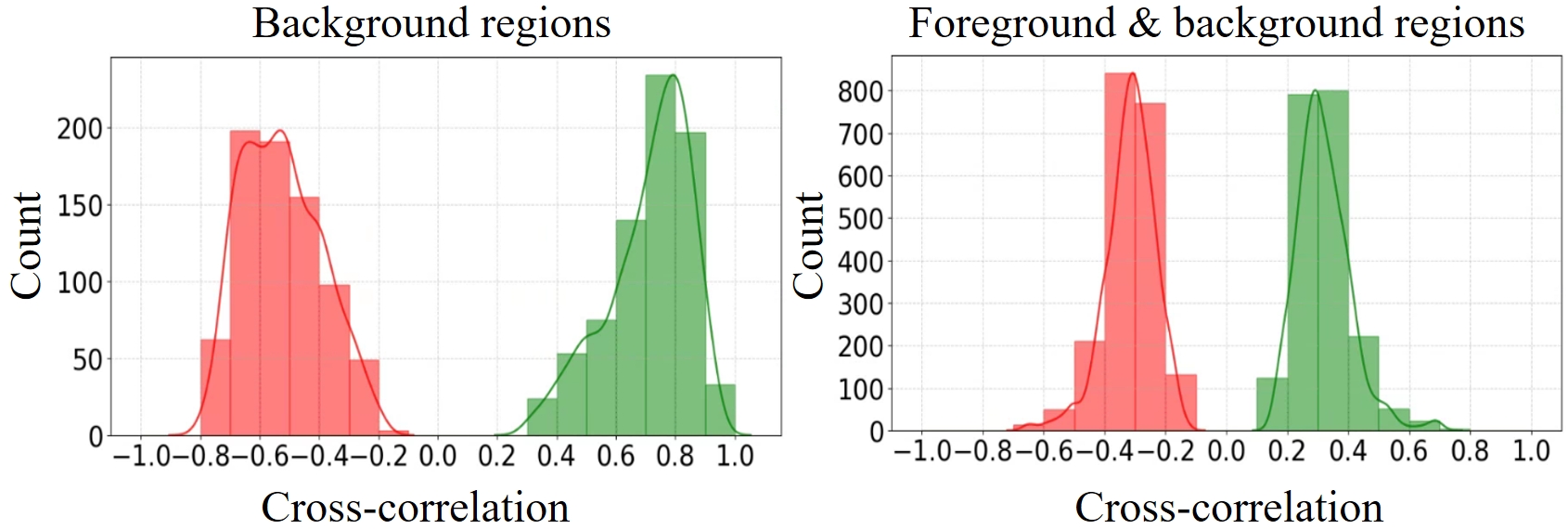}
        \end{minipage} \\ (a)
        \\        
        \begin{minipage}{.48\textwidth}
            \includegraphics[width=\linewidth]{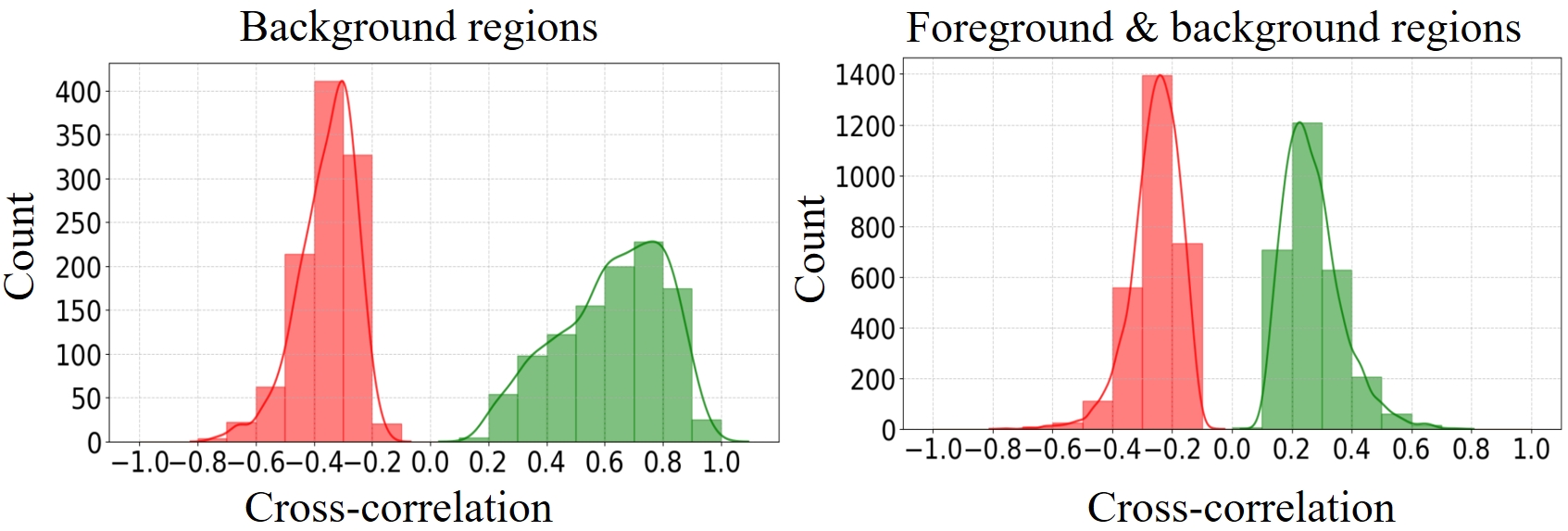}
        \end{minipage} \\ (b)
        \\
        \begin{minipage}{.48\textwidth}
            \includegraphics[width=\linewidth]{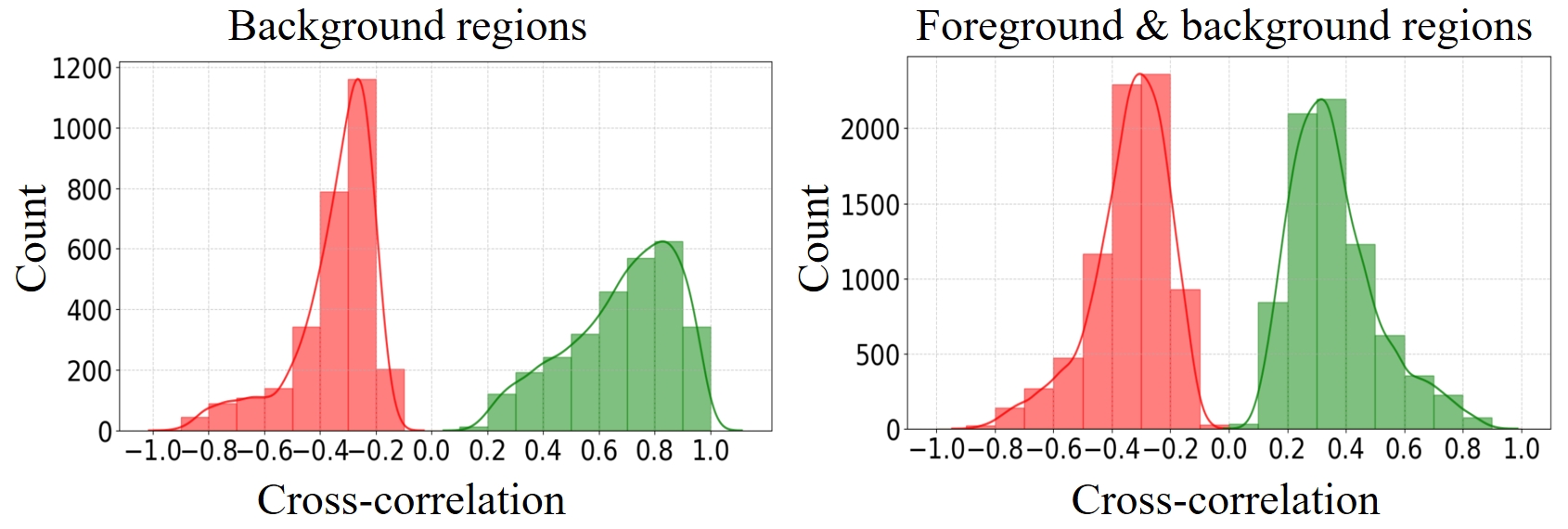}
        \end{minipage} \\ (c)
        \\
         \begin{minipage}{.48\textwidth}
            \includegraphics[width=\linewidth]{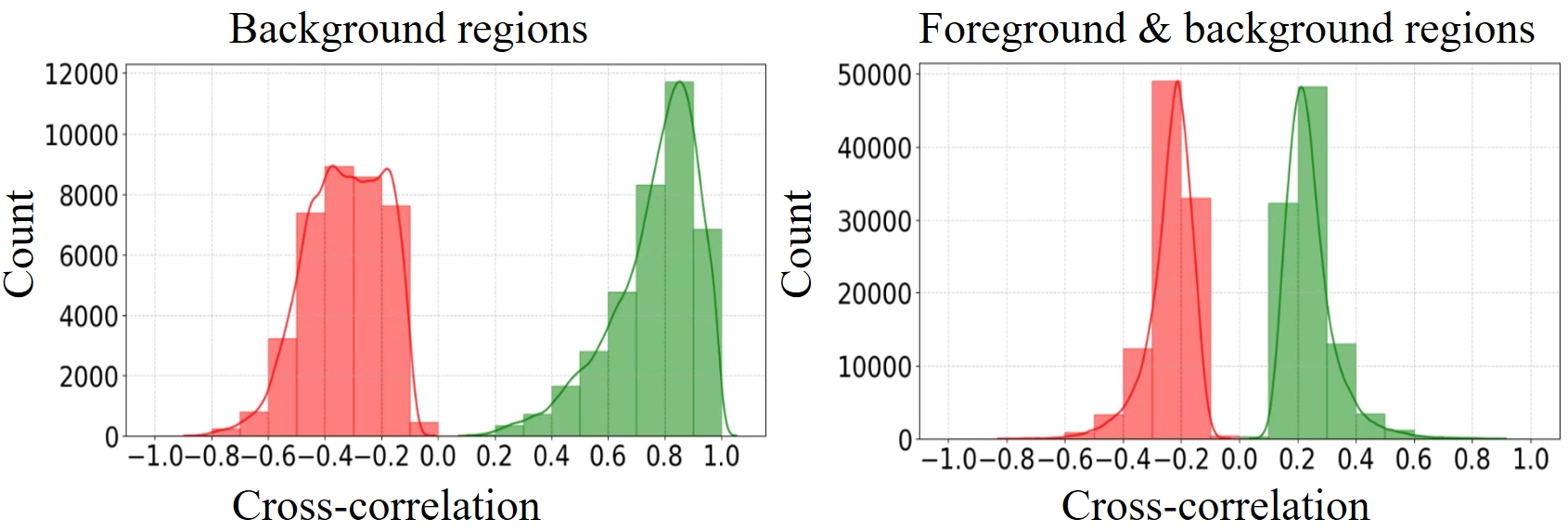}
        \end{minipage} \\ (d)
        \\
          \begin{minipage}{.48\textwidth}
            \includegraphics[width=\linewidth]{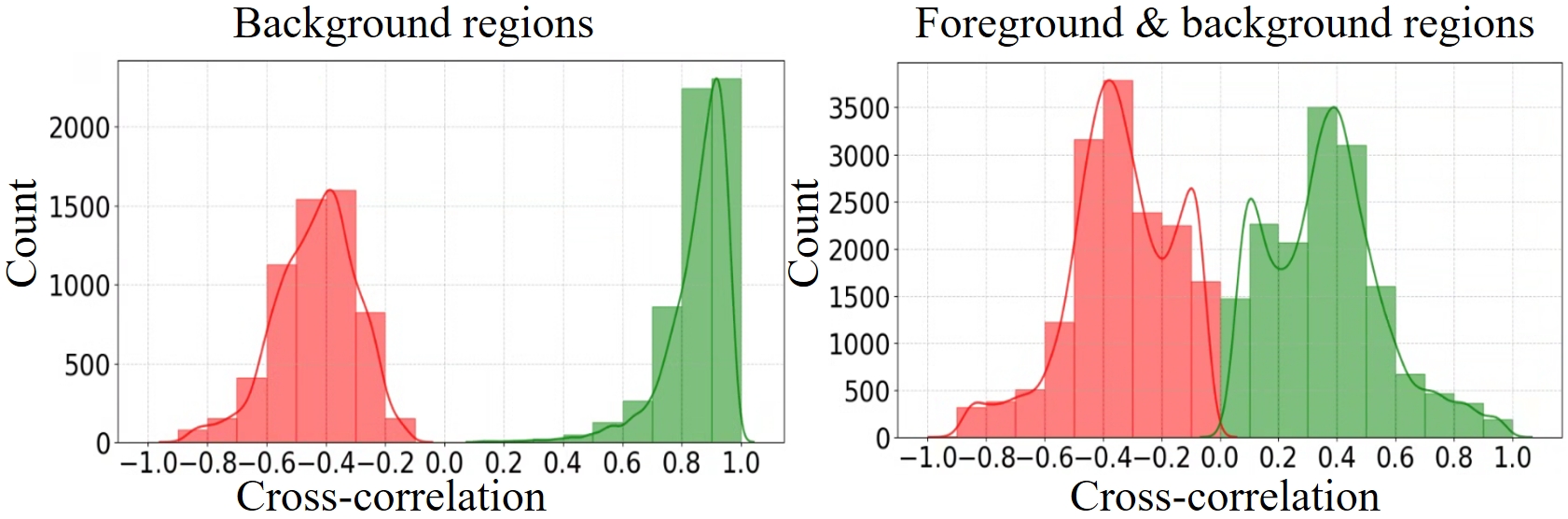}
        \end{minipage} \\ (e)
        \\
    \end{tabular}
\caption{   
Distributions of cross-correlation between signals: (left) extracted from different background regions and (right) extracted from both foreground and background regions, with positive cross-correlation shown in green and negative cross-correlation shown in red, across multiple rPPG datasets: 
(a) \textbf{UBFC-rPPG} \cite{bobbia2019unsupervised}, (b) \textbf{PURE} \cite{stricker2014non}, (c) \textbf{COHFACE} \cite{heusch2017reproducible},(d) \textbf{VIPL-HR} \cite{niu2019rhythmnet}, and (e) \textbf{MR-NIRP Car} \cite{NowaraDriving}.  
} 
\label{fig:local similarity in dataset}  
\end{figure}

\paragraph{ Across Multiple rPPG Datasets}
 
We further conduct experiments to evaluate the local spatial and temporal characteristics of interference across multiple rPPG datasets, including  \textbf{UBFC-rPPG} \cite{bobbia2019unsupervised}, \textbf{PURE} \cite{stricker2014non},  \textbf{COHFACE} \cite{heusch2017reproducible}, \textbf{VIPL-HR} \cite{niu2019rhythmnet}, and \textbf{MR-NIRP Car} \cite{NowaraDriving}.
For each video $x$ in these datasets, we first generate a set of facial foreground clips $C^{fg}$ and a set of non-facial background clips $C^{bg}$ from different foreground and background regions, as detailed later in Sec. IV.B and illustrated in Figure~\ref{fig:preprocessing}.
Next, we again utilize the off-the-shelf rPPG estimator \cite{sun2022contrast} to extract signals from $C^{fg}$ and $C^{bg}$ and compute the running correlations between (1) signals extracted from different non-facial background regions and (2) signals extracted from facial foreground and non-facial background regions. 
As shown in Figure~\ref{fig:local similarity in dataset}, interference signals extracted from different background clips exhibit both strong positive and strong negative correlations across different datasets, indicating that interference similarities exist within local spatial-temporal regions. This phenomenon is especially evident in the \textbf{MR-NIRP Car} \cite{NowaraDriving} dataset, which was captured under non-uniform lighting and dynamically changing backgrounds.
Furthermore, this local characteristic is also observed between interference signals extracted from non-facial background clips and rPPG signals extracted from facial foreground clips. This suggests that the extracted rPPG signals are corrupted by interferences (as discussed in Sec. III.B) and that the local spatial-temporal characteristics of interferences are present not only in non-facial background regions but also in foreground regions.

\begin{table}[t] 
\caption{ 
Similar heart rates (HRs) between the original facial video $x^{fg}$ and its augmented version $aug$-$x^{fg}$. }
\centering
\scriptsize 
\setlength{\tabcolsep}{1pt}
\label{tab:HR differences} 
\begin{tabular}{|c|c|c|c|}
\hline
 Method &  Type & Dataset & \begin{tabular}[c] {@{}l@{}}Average HR differences \\ between $x^{fg}$ and $aug$-$x^{fg}$ \end{tabular}  \\ \hline
\multirow{2}{*}{Contrast-Phys \cite{sun2022contrast}}    &  \multirow{2}{*}{\begin{tabular}[c] {@{}l@{}}Learning-based \end{tabular}}   &  \textbf{ UBFC-rPPG} & 0.49±1.85 \\ 

 &   &  \textbf{PURE} &  0.11±0.8 \\  \hline

 \multirow{2}{*}{CHROM \cite{de2013robust}}    &  \multirow{2}{*}{\begin{tabular}[c] {@{}l@{}}Non-learning \end{tabular}}   &  \textbf{ UBFC-rPPG} & 0.0±0.0  \\ 

 &   &  \textbf{PURE} &  0.0±0.0  \\  \hline

 \multirow{2}{*}{ POS \cite{wang2016algorithmic}}    &  \multirow{2}{*}{Non-learning}   &  \textbf{ UBFC-rPPG} & 0.0±0.0 \\ 

 &   &  \textbf{PURE} &  0.0±0.0  \\  \hline
 
\end{tabular}
\end{table}

\subsection{Characteristics of rPPG Signals}
\label{sec:Characteristics of rPPG}

\subsubsection{Observations of rPPG Signals in Previous Methods}

Previous unsupervised rPPG estimation methods \cite{gideon2021way,sun2022contrast,speth2023non,yue2023facial} have explored and validated several observations about characteristics of rPPG, including HR range constraint, spatial-temporal similarities, and dissimilarity across different videos.

First, as noted in \cite{gideon2021way}, the majority of human heart rates typically fall within the range of 40 to 250 beats per minute (bpm). Accordingly, the authors in \cite{gideon2021way} recommended utilizing this HR range as a constraint for determining rPPG signals with power spectrum densities (PSDs) falling between 0.66 Hz and 4.16 Hz.

Next, in \cite{gideon2021way}, the authors also observed that HR does not change rapidly in a short time and that HR detected from different facial regions should be consistent. This observation implies that rPPG signals exhibit spatial-temporal similarity within a single facial video.

Moreover, in \cite{sun2022contrast}, the authors observed that rPPG signals from different subjects have distinct waveforms and PSDs.
Therefore, they adopted this cross-video dissimilarity within the framework of contrastive learning to improve rPPG estimation.

\subsubsection{Consistency in rPPG with Augmentation}
\label{sec:Consistency of rPPG}

\begin{figure} [t]
 \centering
   \begin{minipage}[b]{0.8\linewidth} 
    \includegraphics[width=7cm]{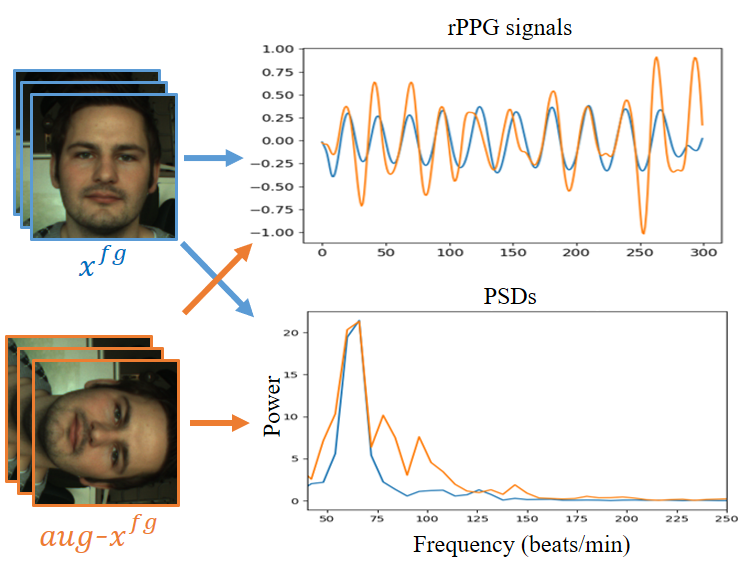} 
  \end{minipage}
  
\caption{
Consistency in rPPG with augmentation: the original facial video $x^{fg}$ and its rotated version $aug$-$x^{fg}$ have similar rPPG waveforms and PSDs.
} 

 \label{fig:Characteristics of rPPG}   
\end{figure}

\begin{figure} [t]
 \centering
   \begin{minipage}[b]{0.7\linewidth} 
    \includegraphics[width=5.cm]{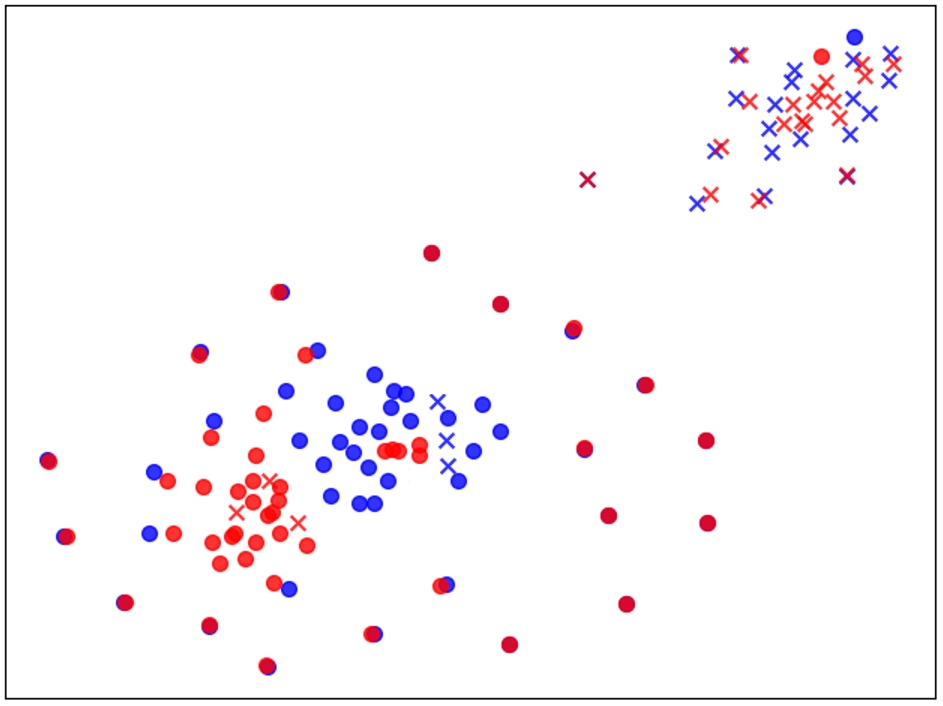} 
  \end{minipage}
\caption{ 
 $t$-SNE visualizations on \textbf{ VIPL-HR} \cite{niu2019rhythmnet} across different HR ranges, including DOTS: 50-60 HRs and CROSS: 90-100 HRs (red: rPPG features from the original facial video $x^{fg}$, blue: rPPG features from the augmented version $aug$-$x^{fg}$). 
 } 

 \label{fig:different rPPG features}   
\end{figure}

Since data augmentation has been popularly adopted to improve model training, we will also incorporate data augmentation in the proposed method. 
In particular, we expect that a facial video and its weakly augmented counterpart through, e.g., rotation, random crop, and flip, should exhibit identical rPPG signals. 
Figure \ref{fig:Characteristics of rPPG} shows an example that the extracted rPPG signals from a facial video and its rotated version indeed have similar waveforms and PSDs.  
In Table~\ref{tab:HR differences}, we measure the average HR differences between the original facial video $x^{fg}$ and its augmented version $aug$-$x^{fg}$ using  both a learning-based method (Contrast-Phys \cite{sun2022contrast}) and non-learning-based methods (CHROM \cite{de2013robust} and POS \cite{wang2016algorithmic}).
The learning-based method,  Contrast-Phys \cite{sun2022contrast}, was trained on datasets different from the two test datasets: \textbf{UBFC-rPPG} and \textbf{PURE}. 
The results in Table~\ref{tab:HR differences} demonstrate that the original and augmented facial videos yield nearly identical heart rates.
Note that, slight HR differences are observed in the learning-based Contrast-Phys \cite{sun2022contrast} method. This HR discrepancy between $x^{fg}$ and $aug$-$x^{fg}$ arises because Contrast-Phys \cite{sun2022contrast} was not trained with rotation augmentation. 
In contrast, the two non-learning based methods, CHROM \cite{de2013robust} and POS \cite{wang2016algorithmic}, which rely on mathematical models to track color changes associated with blood flow, remain unaffected by weak augmentation operations and consistently produce identical HR values for $x^{fg}$ and $aug$-$x^{fg}$. 
In addition, a $t$-SNE visualization of rPPG features from the original and augmented facial videos across different HR ranges, presented later in Figure~\ref{fig:different rPPG features}, further supports this observation. This visualization also aligns with the conclusion in \cite{yang2023simper}  that rPPG features corresponding to similar HRs tend to cluster closely in the latent feature space. 
Therefore, in addition to the rPPG characteristics mentioned in the previous subsection, we will also enforce the consistency in augmentation to constrain the proposed model.

\section{Proposed Method}
\label{sec:Proposed_method}

\begin{figure*} [t]
    \begin{minipage}[b]{1.0\linewidth}
      \centering{ 
      \includegraphics[width=14.5cm]{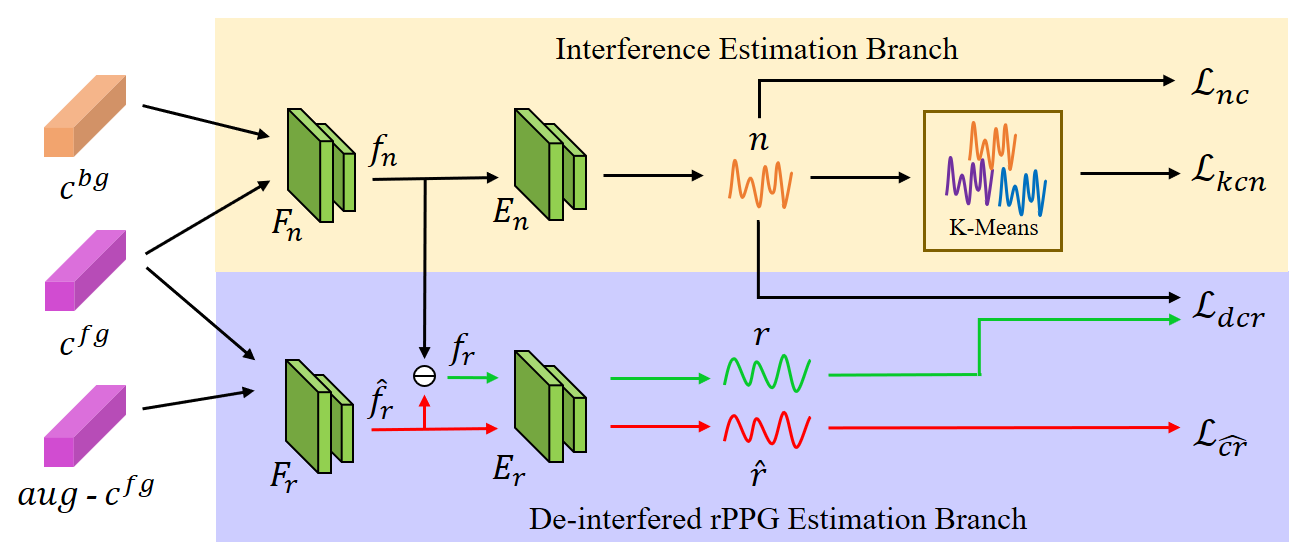}
      }
    \end{minipage} 
\caption{ 
We propose an unsupervised rPPG estimation model: De-interfered and Descriptive rPPG Estimation Network (DD-rPPGNet), to directly learn rPPG signals $r$ from one input background clip $c^{bg}$, one input facial clips $c^{fg}$, and the weakly-augmented facial clip $aug$-$c^{fg}$. 
DD-rPPGNet consists of two branches. 
In the Interference Estimation Branch, we propose to learn interference feature $f_n$ by constraining local spatial-temporal similarity within interference signals $n$. In the De-interfered rPPG Estimation Branch, we propose to learn initial estimates $\hat{f}_{r}$ and $\hat{r}$ of rPPG by enforcing consistency between 
$c^{fg}$ and $aug$-$c^{fg}$, and next to refine the initial rPPG estimates through interference cancellation from $\hat{f}_{r}$ to derive the de-interfered rPPG features $f_r$ and de-interfered rPPG signals $r$. 
} 

 \label{fig:framework}   
\end{figure*}

In this paper, we propose a novel unsupervised rPPG estimation model, called De-interfered and Descriptive rPPG Estimation Network (DD-rPPGNet), to eliminate interference in rPPG estimation without referring to ground truth rPPG signals.
In Section \ref{sec:Overview}, we first give an overview of the proposed DD-rPPGNet. 
Then, in Section \ref{sec:Interference Modeling}, we present how we adopt the local spatial-temporal characteristics to model the interference signals.
Next, in Section \ref{sec:unsupervised de-nosing}, we adopt the rPPG characteristics and also refer to the estimated interference to derive the de-interfered rPPG signals.
Finally, in Section \ref{sec:Learning Discriminative features}, we develop a 3D Learnable Descriptive Convolution (3DLDC) to enhance learning descriptive rPPG features for further improving rPPG estimation.
 
\subsection{Overview of DD-rPPGNet}
\label{sec:Overview} 

Figure \ref{fig:framework} shows the proposed DD-rPPGNet, which consists of two branches: one Interference Estimation Branch and one De-interfered rPPG Estimation Branch. 
First, in the Interference Estimation Branch, we adopt the local spatial-temporal similarity of interference mentioned in Section \ref{sec:Distinguishing Characteristics of Noises} to learn the interference features $f_n$ and to estimate the interference signals $n$. 
Next, in the De-interfered rPPG Estimation Branch,
 we adopt the characteristics of rPPG signals mentioned in Section \ref{sec:Characteristics of rPPG} to learn the inference-carrying rPPG features $\hat{f}_{r}$ and signals $\hat{r}$, and then use $f_n$, $\hat{f}_{r}$, and $n$ to learn the de-interfered rPPG signals $r$.
In addition, we propose an effective 3DLDC by involving learnable descriptors into 3DCNN to capture subtle chrominance changes for learning descriptive rPPG features.

\subsection{Unsupervised Interference Estimation}
\label{sec:Interference Modeling}

In the Interference Estimation Branch, we adopt the local spatial-temporal similarity of interference mentioned in Section \ref{sec:Distinguishing Characteristics of Noises} to model the interference signals $n$ for learning the interference features $f_n$.

First, we describe how we determine the foreground facial component $x^{fg}$ and the background non-facial component $x^{bg}$ from an input video $x$.
As shown in Figure \ref{fig:preprocessing}, we adopt 3DDFA \cite{guo2020towards,zhu2017face} to locate the facial region $x^{fg}$ and then follow the setting described in \cite{yu2021transrppg} to crop both the left and right regions of $x^{fg}$ to generate the non-facial background video $x^{bg}$ with the same size as $x^{fg}$. Next, we randomly sample $L$ clips, each having $\Delta_t$ frames with a fixed size of $h \times w$, from $x^{fg}$ and $x^{bg}$ to generate the input foreground clips $C^{fg} = \{c_l^{fg} \in  \mathbb{R}^{h \times w \times \Delta_t} \}_{l=1}^L$ and the background clips $C^{bg} = \{c_l^{fg} \in  \mathbb{R}^{h \times w \times \Delta_t} \}_{l=1}^L$, respectively.

As shown in Figure \ref{fig:framework}, given the input clips $c^{fg} \in C^{fg}$ and $c^{bg} \in C^{bg}$, our goal in Interference Estimation Branch is to estimate the interference signal $n=E_n ( F_n (c))$ via learning the interference feature extractor $F_n$ and the interference estimator $E_n$.
Assuming $n \approx n^{fg} \approx n^{bg}$, we propose to estimate $n$
by constraining the local spatial-temporal similarity between $n^{fg}$ and $n^{bg}$.
Let $N^{fg}=\{n^{fg}\}$ and $N^{bg}=\{n^{bg}\}$ denote the sets of estimated interference from the input clips $C^{fg}=\{c^{fg}\}$ and $C^{bg}=\{c^{bg}\}$ within a batch, respectively.
We first define the interference correlation loss $\mathcal{L}_{nc}$ to enforce the local spatial-temporal similarity between $n^{fg}$ and $n^{bg}$ by,
\begin{normalsize}
\begin{eqnarray}  
\mathcal{L}_{nc} = \mathbb{E} \{ NP( {n}^{fg} , {n}^{bg}) \},
\label{eq:sim}  
\end{eqnarray}
\end{normalsize}

\noindent
where $\mathbb{E}\{\cdot\}$ denotes the expectation,
 $NP(  {n}^{fg} , {n}^{bg} ) = 1- \frac{Cov(  {n}^{fg} , {n}^{bg} )}{\sqrt{Cov(  {n}^{fg} ,  {n}^{fg} )}\sqrt{Cov( {n}^{bg} , {n}^{bg} )}} $ is the negative Pearson correlation  \cite{tsou2020multi, tsou2020siamese, hu2021eta, lu2021dual} between $n^{fg}$ and $n^{bg}$, 
 and $Cov({n}^{fg} , {n}^{bg})$ denotes  the covariance between $n^{fg}$ and $n^{bg}$.

In addition, to further enforce the local spatial-temporal similarity of interference, we adopt $\operatorname{K-Means}(\cdot)$ \cite{yadav2013review, gunecs2010efficient, xu2014cluster} to aggregate the estimated interference signals into $K$ clusters ${N}_{k}, k=1,…,K$. Next, we define the K-Means-based contrastive interference loss $\mathcal{L}_{kcn}$ to pull together similar signals from the same cluster and to push away relatively  dissimilar signals from other clusters by,

\begin{normalsize} 
\begin{equation}
\begin{split}
\mathcal{L}_{kcn} = \mathbb{E}\{\log (\frac{\exp(NP(n_i,n_j) }
{ \sum\limits_{\substack{n_h \in  N^{fg} \cup N^{bg}}}
\exp(NP(n_i,n_h) )} + 1) \} ,
\label{eq:corruption_contrastive}  
\end{split}
\end{equation}
\end{normalsize}
 
\noindent  
where $n_i, n_j \in N_k$ are the estimated interference signals from the same clusters, and
$n_h \in  N^{fg} \cup N^{bg}$ denotes all the estimated interference signals.
   
\subsection{Unsupervised De-interfered rPPG Estimation}
\label{sec:unsupervised de-nosing}

In the De-interfered rPPG Estimation Branch, we leverage the characteristics of rPPG signals described in Section \ref{sec:Characteristics of rPPG} and refer to the interference features $f_n$ extracted in the Interference Estimation Branch to directly estimate the rPPG signals without relying on the ground truth.

First, we conduct weak augmentation, including rotation and flip operations, on the original facial clips $C^{fg}$ to generate augmented video clips $aug$-$C^{fg}$ as an additional input to this branch. From the characteristics of rPPG consistency with augmentation, our goal is to derive the initial estimates of rPPG features $ \hat{f}_r=F_r(c)$ and rPPG signals $ \hat{r} = E_r(\hat{f}_r)$ by learning the rPPG feature extractor $F_r$ and the rPPG estimator $E_r$. 
To achieve this goal, we define the contrastive rPPG loss $\mathcal{L}_{\widehat{cr}}$ to pull together the rPPG signals $\hat{r}$ extracted from a video clip and its augmented counterpart while simultaneously pushing away the rPPG signals $\hat{r}$ extracted from different videos by,

\begin{normalsize} 
\begin{equation}
\begin{split} 
\mathcal{L}_{\widehat{cr}} = \mathbb{E}\{
\log( \frac{\exp(d(P(\hat{r}_i),P(\hat{r}_j) ) }
{   \sum \limits_{
\substack{\hat{r}_h \in \hat{R}_{all}} 
}  \exp(d(P(\hat{r}_i),P(\hat{r}_h)) )} +  1 )  \} ,
\label{eq:unsupervised_contrastive}  
\end{split}
\end{equation}
\end{normalsize}  

\noindent
where $\hat{r}_i,\hat{r}_j \in \hat{R}_b$ denote the estimated rPPG signals from $C_b^{fg}$ and its augmented counterpart $aug$-$C_b^{fg}$ of one single video $x^{fg}_b$, respectively; $P(\cdot)$ denotes  the computation of PDS, $d(\cdot)$ measures the Euclidean distance, and $\hat{r}_h \in \hat{R}_{all}$  denotes the estimated rPPG signals from all the videos and their augmented counterparts within one batch.

Incorporating with the Interference Estimation Branch, our next goal is to extract de-interfered rPPG features $f_r$ by canceling the interference features $f_n$ from the initial estimates $\hat{f}_r$ so as to derive de-interfered rPPG signal by $r = E_r(f_r)$. 
Our inspiration comes from a recent supervised rPPG method \cite{du2023dual}, which proposed to learn de-noised rPPG features through a residual method by subtracting noise features from the noise-carrying rPPG features.
To expand upon this concept in our unsupervised learning model, we propose to derive de-interfered rPPG features by ${f}_{r} = {\hat{f}}_{r} - {f}_{n}^{fg}$, where ${f}_{n}^{fg} = F_n (c^{fg})$ is the interference features extracted from the input facial clips. After obtaining the de-interfered rPPG features ${f}_{r}$, we next derive a new set of rPPG signals ${R} = \{ {r}_i \}  $ from $C^{fg}\cup aug$-$C^{fg}$, where $r = E_r(f_r)$ is the refined estimate of rPPG signals.

Subsequently, similar to Equation \eqref{eq:unsupervised_contrastive}, we once again adopt the idea of contrastive learning by replacing the set of initial estimates $\hat{R}$ with the set of refined estimates $R$.

Furthermore, we additionally encourage the refined estimates $r$ to be away from the interference $n$ and define the de-interfered contrastive rPPG loss $\mathcal{L}_{dcr}$ by,
\begin{footnotesize} 
\begin{equation}  
\begin{split}
\mathcal{L}_{dcr} \nk\nk = \nk \mathbb{E}\{
 \nk \log( \nk\frac{\nk\nk\exp(d(P(r_i),P(r_j))\nk)}
{ \sum \limits_{
\substack{r_h  \in \\ \nk {R}_{all}} 
} \nk\nk \exp(\nk d\nk (P(r_i), P(r_h)) \nk ) \nk\nk + \nk\nk 
\sum\limits_{\substack{n_h \in \\ N^{fg}}} \nk\nk \exp(d \nk (P(r_i),  P(n_h))\nk )} \nk\nk + \nk 1 \nk) \nk\},
\label{eq:unsupervised_contrastive_dcr}  
\end{split}
\end{equation}
\end{footnotesize}

\noindent
where $r_i,r_j\in R_b$ are the refined rPPG signals estimated from $C_b^{fg}$ and $aug$-$C_b^{fg}$ of the input video $x^{fg}_b$, respectively; $P(\cdot)$ denotes  the computation of PDS, $d(\cdot)$ measures the Euclidean distance, 
$n_h \in N^{fg}$ denotes the estimated interference from $C_b^{fg}$, and 
$r_h \in R_{all}$  denotes the refined rPPG signals estimated from all the videos and their augmented counterparts within one batch.

\begin{figure}
    \centering
    \begin{tabular}{cc} 
    {\includegraphics[height=6.5cm]{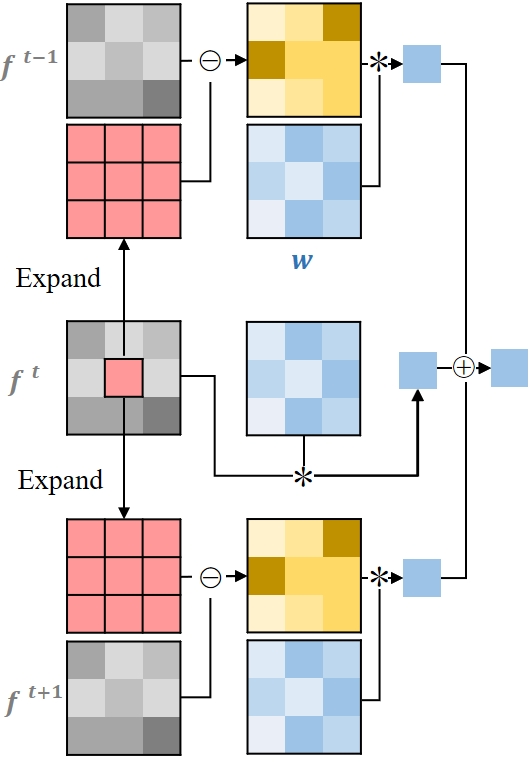}} &
    {\includegraphics[height=6.5cm]{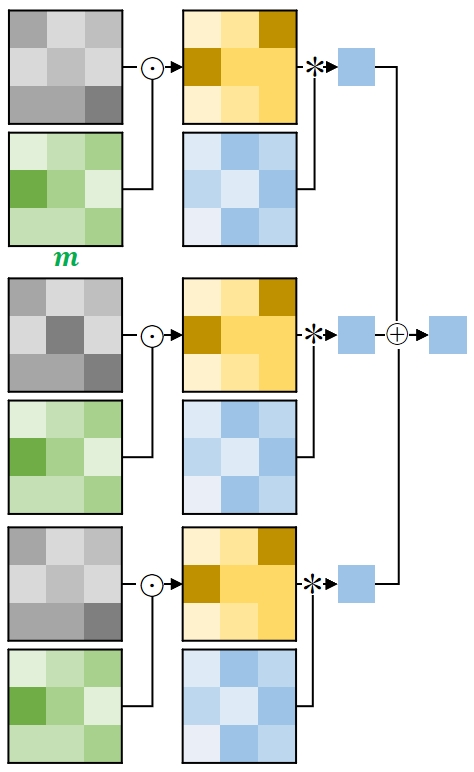}} 
    \\ (a)&(b) 
    \end{tabular}
\caption{
Illustration of (a) Temporal Difference Convolution  (TDC) \cite{yu2020autohr} and (b) the proposed 3DLDC.
$\odot$, $*$ and $\ominus$ denote the element-wise multiplication, the convolution operation and the subtraction, respectively.
} 
\label{fig:TDC_3DLDC} 
\end{figure}

\subsection{ Descriptive rPPG Feature Learning}
\label{sec:Learning Discriminative features}

To extend the description capability of CNN and Transformer, most previous rPPG estimation methods \cite{yu2020autohr,yu2022physformer,zhao2021video} adopted predefined descriptors, such as 3D Central Difference Convolution (3DCDC) \cite{zhao2021video} in CNN, to capture 
 the temporal differences of gradient information.

Because these predefined descriptors \cite{yu2020autohr,yu2022physformer,zhao2021video} cannot well adapt to subtle changes along temporal domain on uneven skin caused by varying reflection changes \cite{wong2022optimising}, we believe including a learnable descriptor into 3D convolution
would positively enhance the rPPG feature descriptive ability.
To enable the model to adaptively focus on blood volume changes in optical information, we extend our previously proposed 2DLDC in \cite{huang2022learnable} and propose a novel 3D Learnable Descriptive Convolution (3DLDC) to learn descriptive rPPG features. 

\subsubsection{2D Learnable Descriptive Convolution and 3D Temporal Difference Convolution}
\label{sec:2DLDC and 3D TDC}

We first recap our previous design of 2DLDC \cite{huang2022learnable} and the operation of 3D Temporal Difference Convolution \cite{yu2020autohr,yu2022physformer}. As mentioned in \cite{huang2022learnable}, predefined and unlearnable descriptors are inflexible to capture various textural changes across different materials and limit the representation capacity of vanilla 2D convolution.
Therefore, in 2D Learnable Descriptive Convolution (2DLDC) \cite{huang2022learnable,huang2023ldcformer}, we included a learnable local descriptor into vanilla 2D convolution to adaptively capture subtle textures on different materials and thus increase its representation capacity. 
The design of 2DLDC \cite{huang2022learnable} is to incorporate a learnable descriptor $m$ into vanilla convolution by,

\begin{normalsize}  
\begin{eqnarray} 
\label{eq:learnable_convolution}
\begin{split} 
g(p^t) =& (1-\epsilon)\underbrace{ \sum\limits_{p_k^t \in \mathcal{R}^t}  w(p_k^t) \cdot f(p^t + p_k^t)}_\text{vanilla \;  convolution} \\ 
&+ \epsilon \underbrace{ \sum\limits_{p_k^t \in \mathcal{R}^t}  w(p_k^t) \cdot ( f(p^t + p_k^t) 
\cdot m(p_k^t))}_\text{learnable \; descriptive \; convolution},
\end{split}
\end{eqnarray}
\end{normalsize} 

\noindent 
where $w$ is the convolution kernel, $f$ is the input feature map, $p^t$ is the pixel of current location in the $t$-th frame, $p_k^t$ is the location of neighboring pixels in a local neighborhood $\mathcal{R}^t = \{(-1,-1),(-1,0),...,(0,1),(1,1)\}$, $g$ is the output feature map, and the hyperparameter $\epsilon \in$  [0, 1] 
accounts for tradeoffs between
vanilla and 2D LDC convolution.

Although 3D convolution (3DCNN) has been widely employed in rPPG estimation task, vanilla 3DCNN often leads to poor performance under challenging conditions due to its limited representation capacity.
Therefore, 
the authors in \cite{yu2020autohr}  proposed Temporal Difference Convolution (TDC) by adopting a predefined descriptor to involve the temporal differences information into 3DCNN to provide fine-grained temporal context for robust rPPG estimation. 
As shown in Figure \ref{fig:TDC_3DLDC} (a), TDC aggregates the  temporal difference clues within local temporal regions  $\mathcal{R}^{t-1}$, $\mathcal{R}^{t}$, and $\mathcal{R}^{t+1}$ on the feature level by,
\begin{equation}
\begin{split}
g(p^t) \nk=\nk& \sum_{p^{t\nk-\nk1}_k \in \mathcal{R}^{t\nk-\nk1}} 
w(p_k^{t-1}) \nk\cdot\nk (f(p^{t-1} \nk+\nk p_k^{t-1}) \nk-\nk \theta \nk\cdot\nk f(p^t)) \\
&\nk+\nk\sum_{p^{t\nk+\nk1}_k \in \mathcal{R}^{t\nk+\nk1}} 
w(p_k^{t+1}) \nk\cdot\nk (f(p^{t+1} \nk+\nk p_k^{t+1}) \nk-\nk \theta \nk\cdot\nk f(p^t)) \\
&\nk+\nk\sum_{p^t_k \in \mathcal{R}^t} 
w(p_k^t) \nk\cdot\nk f(p^t \nk+\nk p_k^t),
\label{eq:TDC}  
\end{split}
\end{equation} 

\noindent
where the hyperparameter $\theta \in$  [0, 1] specifies the contribution of temporal difference.

\subsubsection{3D Learnable Descriptive Convolution}
\label{sec:3DLDC}

To effectively learn descriptive rPPG features, as shown in Figure \ref{fig:TDC_3DLDC} (b), we extend our previously proposed 2DLDC \cite{huang2022learnable}  and  propose an effective 3D Learnable Descriptive Convolution (3DLDC)   by  incorporating a learnable descriptor $m$ into vanilla 3D convolution to capture subtle skin color changes by,

\begin{equation}
\begin{split}
g(p^t)
\nk= & (1 \nk-\nk \epsilon) \underbrace{\sum_{i} \nk \sum_{p^{t\nk+\nk i}_{k} \in \mathcal{R}^{t\nk+\nk i}} w(p_k^{t\nk+\nk i}) \nk\cdot\nk f(p^{t\nk+\nk i} \nk+\nk p_k^{t\nk+\nk i})}_\text{vanilla \;  convolution} \\
&\nk+\nk\epsilon \underbrace{\sum_{i} \nk\nk \sum_{p^{t \nk+\nk i}_{k} \nk\in\nk \mathcal{R}^{t\nk+\nk i}} \nk\nk\nk w(p_k^{t\nk+\nk i}) \nk\nk\cdot\nk\nk (f(p^{t\nk+\nk i} \nk\nk+\nk\nk p_k^{t\nk+\nk i}) \nk\nk\cdot\nk m(p_k^{t\nk+\nk i})\nk)}_\text{3D \; learnable \; descriptive \; convolution},    
\label{eq:3DLDC}  
\end{split} 
\end{equation}  

\noindent 
where the learnable descriptor $m$ and the convolution kernel $w$ are both of the same size $3 \times 3 \times 3$ and $i \in \{-1,0,1\}$.

Note that, the proposed 3DLDC exhibits a good generalization of other convolutions. 
When $\epsilon=0$, 3DLDC apparently becomes vanilla 3D convolution. 
In addition,  by comparing Equation  \eqref{eq:TDC} with Equation  \eqref{eq:3DLDC}, we show that Temporal Difference Convolution (TDC) \cite{yu2020autohr} is a special case of 3DLDC when  the matrix $m$ in Equation \eqref{eq:3DLDC} is  

\begin{normalsize} 
\begin{eqnarray}
\begin{split} 
m = \textbf{1}_{3 \times 3 \times 3} + 
  \left[ { 
      m_0,
      \left[ {\begin{array}{ccc}
        0 & 0 & 0 \\
        0 & w_s& 0 \\
        0 & 0 & 0 
      \end{array} } \right],
       m_0
} \right], 
\label{eq:LDC_special_case}
 \end{split}
\end{eqnarray}
\end{normalsize}

\noindent
where
\begin{normalsize}
\begin{eqnarray}
\begin{split}  
w_s = -\frac{1}{w(p^t)} (
\sum_{p^{t-1}_{k} \in \mathcal{R}^{t-1}} w(p^{t-1}_k)+\sum_{p^{t+1}_{k} \in \mathcal{R}^{t+1}} w(p^{t+1}_k))
\label{eq:LDC_special_case_weight}
 \end{split}
\end{eqnarray}
\end{normalsize}

\noindent   
is the weight of the special case in 3DLDC, $m_0$ is the all-zero matrix \textbf{0}$_{3 \times 3}$, and the base matrix $\textbf{1}_{3 \times 3  \times 3}$ is an all-ones matrix. Detailed derivations of Equation \eqref{eq:LDC_special_case} are given in the supplementary material. 
 
\begin{figure} 
    \centering
    \begin{tabular}{c c}
        \begin{minipage}{.4\textwidth}
            \includegraphics[width=\linewidth]{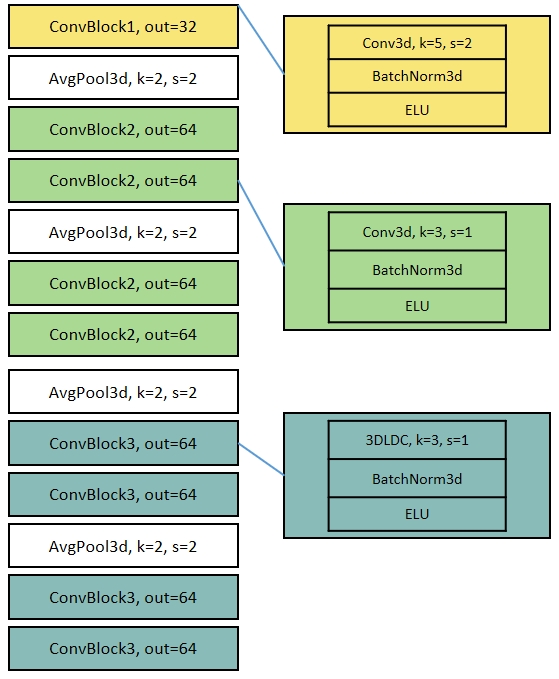}
        \end{minipage} \\ (a)
        \\        
        \begin{minipage}{.4\textwidth}
            \includegraphics[width=\linewidth]{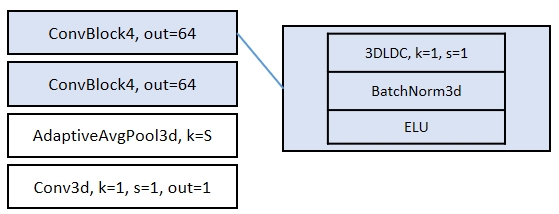}
        \end{minipage} \\ (b)
        \\ 
    \end{tabular}
\caption{   
Network architectures of (a) the interference feature extractor $F_n$ and the rPPG feature extractor $F_r$, and (b) the interference estimator $E_n$ and the rPPG estimator $E_r$. 
}
\label{fig:networks}  
\end{figure}

With the proposed 3DLDC, we replace all vanilla 3DCNNs in the high-level convolution blocks of DD-rPPGNet with 3DLDC. 
The detailed network architectures of the two feature extractors, $F_n$ and $F_r$, as well as the two estimators, $E_n$ and $E_r$, are shown in Figures~\ref{fig:networks} (a) and (b), respectively.

\section{Experiments}
\subsection{Experimental Setting}

\begin{figure} 
    \centering
    \subfloat[]{
    \frame{\includegraphics[width=4.3cm]{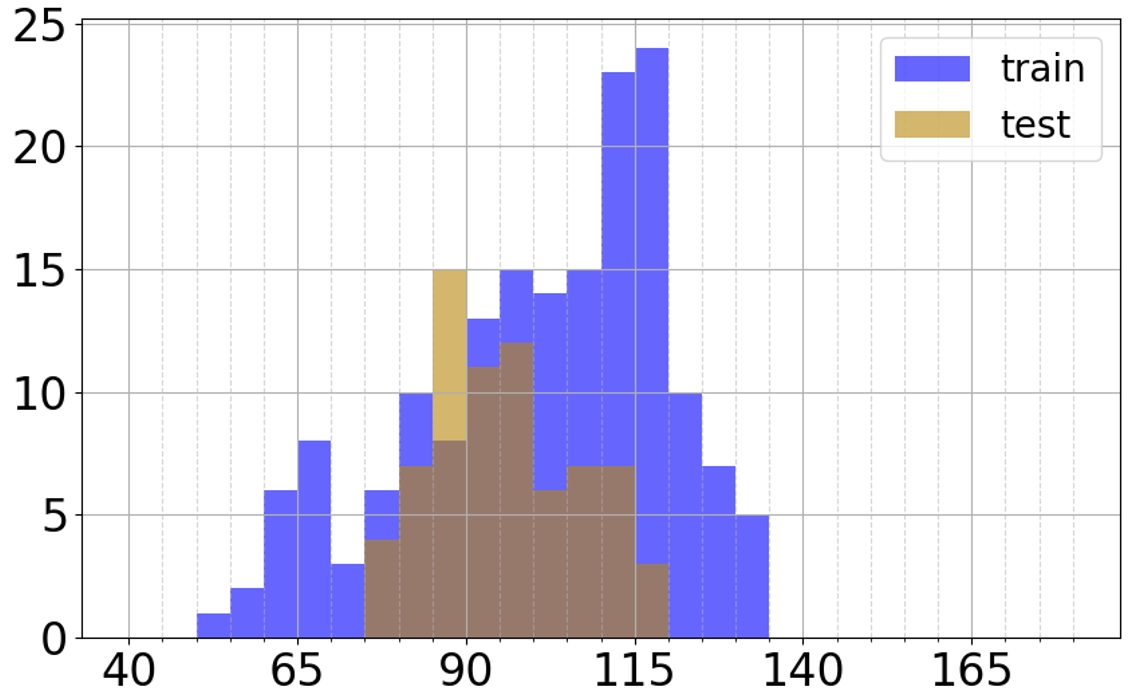}}}
    \hfill
    \subfloat[]{
    \frame{\includegraphics[width=4.3cm]{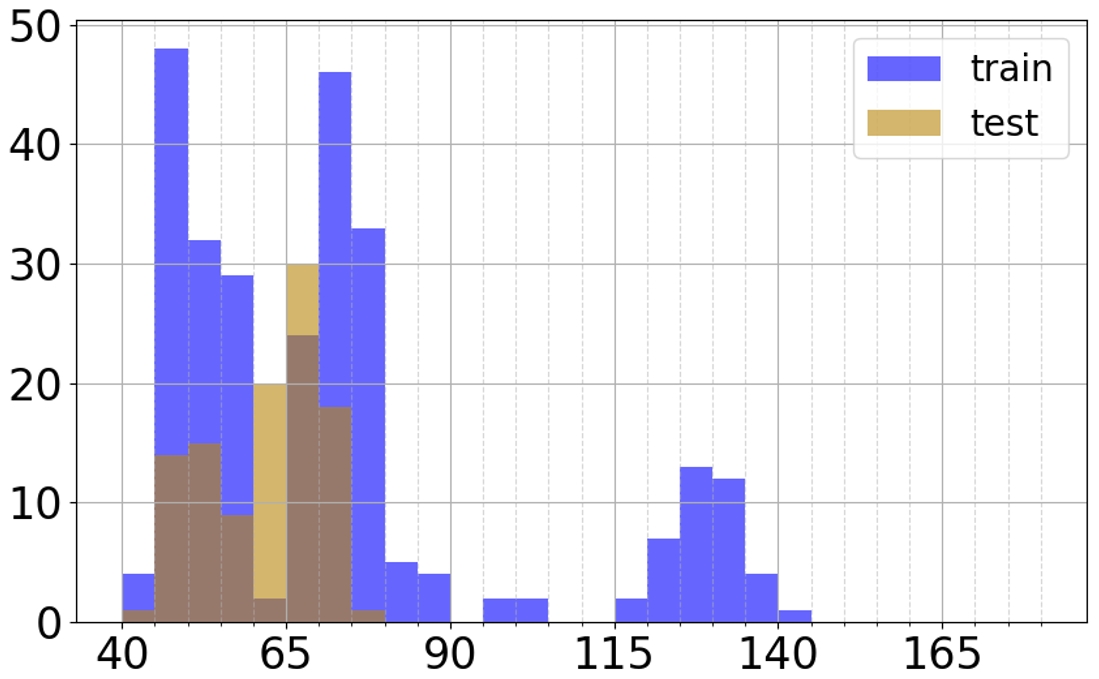}}}
    \vfill
     \subfloat[]{
    \frame{\includegraphics[width=4.3cm]{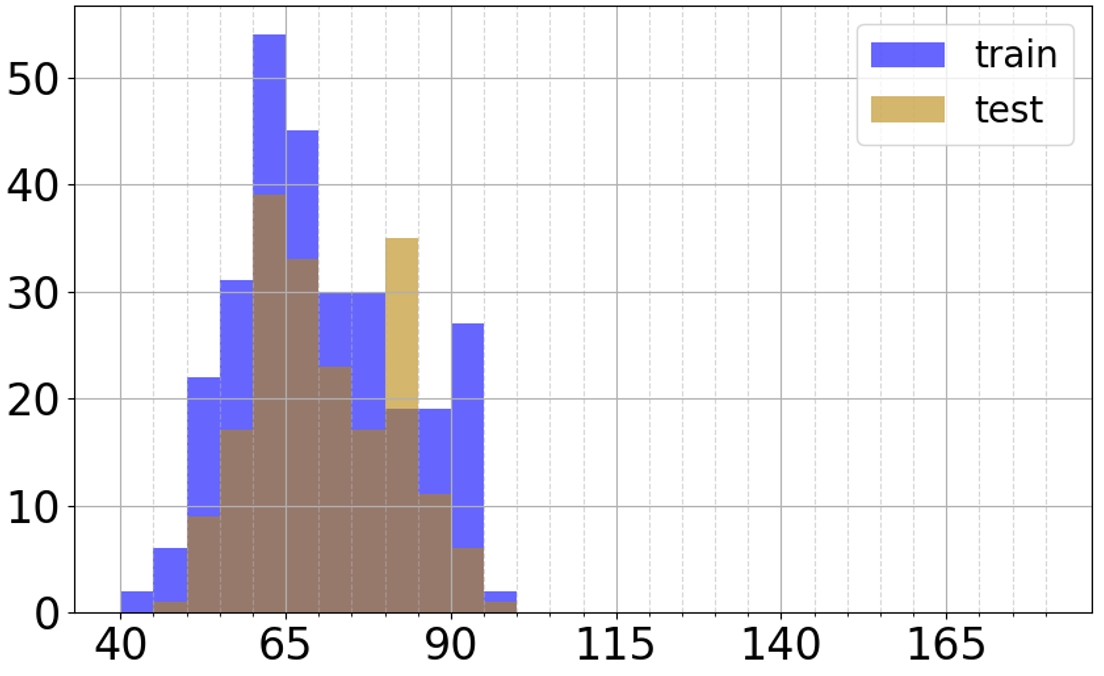}}}
     \hfill
      \subfloat[]{
    \frame{\includegraphics[width=4.3cm]{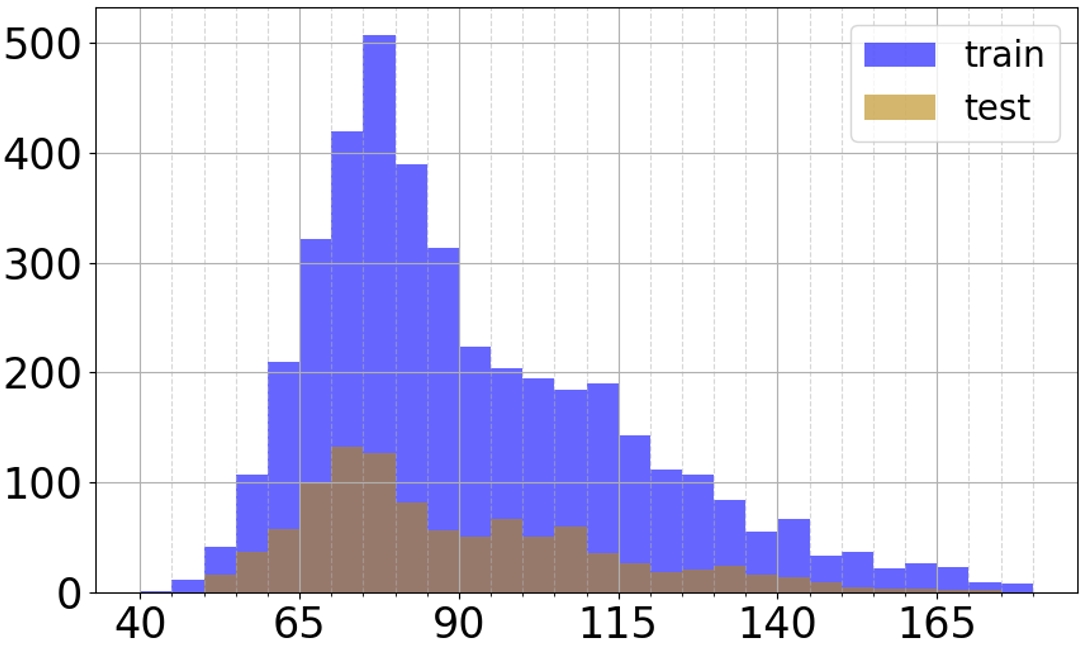}}}
    \vfill
      \subfloat[]{
    \frame{\includegraphics[width=4.3cm]{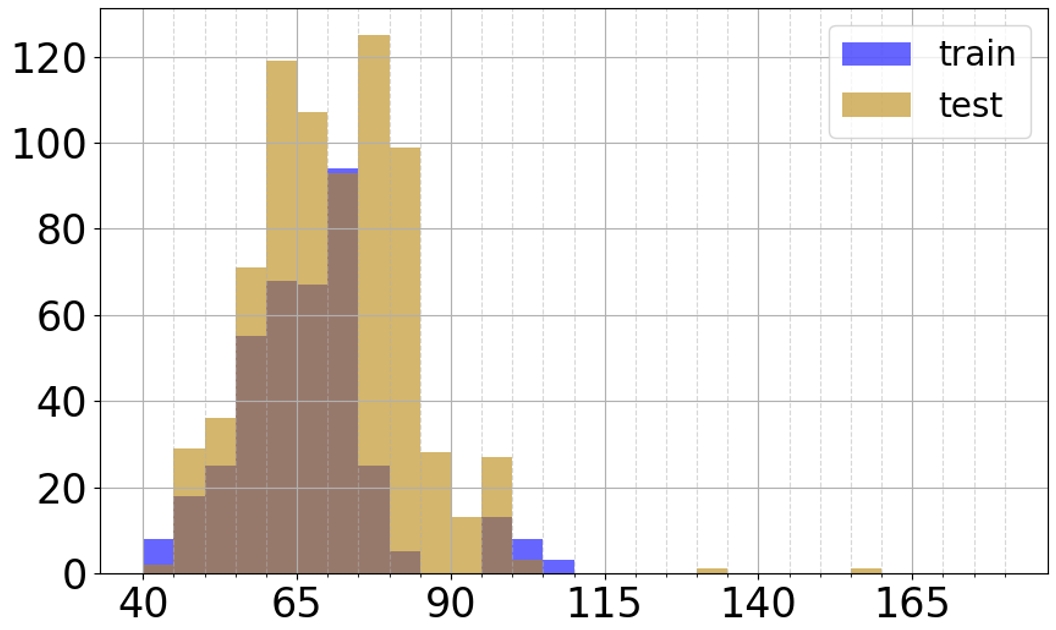}}}
     
\caption{   
Heart rate distribution of the training/testing data split in : (a) \textbf{U} \cite{bobbia2019unsupervised}, (b) \textbf{P} \cite{stricker2014non}, (c) \textbf{C(Natural)} \cite{heusch2017reproducible}, (d) \textbf{V} \cite{niu2019rhythmnet}, and (e) \textbf{Car(RGB)} \cite{NowaraDriving}.
The x-axis represents HR ranges, while the y-axis indicates the number of samples.
} 
\label{fig:distribution in dataset}  
\end{figure}  

\subsubsection{Network architecture}
Figure~\ref{fig:networks} illustrates the detailed network architectures of the proposed DD-rPPGNet. 
As shown in Figure~\ref{fig:networks} (a), both feature extractors, $F_n$ and $F_r$, share the same architecture, consisting of one vanilla convolutional block (ConvBlock1), four vanilla convolutional blocks (ConvBlock2), and four 3DLDC convolutional blocks (ConvBlock3).
Similarly, for the two estimators, $E_n$ and $E_r$, as shown in  Figure~\ref{fig:networks} (b), we adopt the same network architecture, consisting of two 3DLDC convolutional blocks (ConvBlock4).

\subsubsection{Datasets}
We  conduct experiments on the following rPPG databases:  \textbf{ UBFC-rPPG} \cite{bobbia2019unsupervised}  (denoted as \textbf{U}), \textbf{PURE} \cite{stricker2014non} (denoted as \textbf{P}),  \textbf{VIPL-HR} \cite{niu2019rhythmnet} (denoted as \textbf{V}), and \textbf{MR-NIRP Car} \cite{NowaraDriving} (denoted as \textbf{Car}), and \textbf{COHFACE}  \cite{heusch2017reproducible} (denoted as \textbf{C(Natural)} or \textbf{C}). Here, \textbf{C(Natural)} includes only facial videos recorded under ``natural" illumination conditions, while \textbf{C} encompasses facial videos recorded under both ``studio" and ``natural" illumination conditions.
The heart rate distribution in the training and testing data across these databases varies significantly, as shown in Figure~\ref{fig:distribution in dataset}.
For experiments on \textbf{UBFC-rPPG} and \textbf{PURE} databases, we follow the protocol in \cite{hsieh2022augmentation} to train rPPG models on 30 and 7 subjects, respectively, and then test on the remaining 12 and 3 subjects.  
In \textbf{COHFACE}, we train rPPG models on 24 subjects and then test on 12 subjects. In addition, for intra-domain testing, we focus on challenging illumination variations and include only facial videos recorded under ``natural" illumination conditions (see Figure~\ref{fig:ablation_visualization} (a) for an example).
For \textbf{VIPL-HR}, we train on subjects from folds 2, 3, 4, and 5, and test on subjects from fold 1.
In \textbf{MR-NIRP Car}, to focus on challenging environmental variations, we train on videos recorded in a garage under still conditions and test on videos captured during driving with small motion. 

\subsubsection{Evaluation Metrics}
To have a fair comparison with previous methods \cite{chen2018deepphys, yu2020autohr, song2021pulsegan, lu2021dual, nowara2021benefit, hsieh2022augmentation, yu2022physformer, gideon2021way, sun2022contrast, speth2023non, tsou2020multi, chung2022domain},  we follow the same setting in \cite{sun2022contrast} to conduct intra-testing and use the benchmark proposed in \cite{chung2022domain} to conduct cross-testing.
We report all the results in terms of the following evaluation metrics: Mean Absolute Error (MAE), Root Mean Square Error (RMSE), and Pearson Correlation Coefficient (R).

\subsubsection{Implementation Details}  
We implement the proposed DD-rPPGNet by Pytorch and empirically set $P_n= 4$, $P_r = 2$, and $D =4$.  
To train DD-rPPGNet, we set a constant learning rate of 1$\mathbf{e-}$5 with AdamW  \cite{loshchilov2017decoupled} optimizer for 100 epochs.  
The training process is conducted with a batch size of 2 using an NVIDIA GeForce RTX 4090 GPU. 
During the inference stage, we follow previous unsupervised methods \cite{gideon2021way,sun2022contrast} to decompose each test video into non-overlapping 30s clips and use the estimated rPPG signals of each clip to  calculate the power spectrum densities (PSD) for deriving HR of each clip.

\begin{table}[t]
\caption{Ablation study on the protocol of  cross-testing  \textbf{U$+$P$\rightarrow$C}, using different combinations of modules and losses.}
\centering
\footnotesize
\setlength{\tabcolsep}{2pt} 
\label{tab:ablation_different_modules}
\begin{tabular}{cc|cccc|ccc}
\hline
\multicolumn{2}{c|}{Different Modules}           & \multicolumn{4}{c|}{Different Losses}                                                                                                                        & \multicolumn{3}{c}{\textbf{U}$+$ \textbf{P}$\rightarrow$ \textbf{C}} \\ \hline
\multicolumn{1}{c|}{3DCNN}        & 3DLDC        & \multicolumn{1}{c|}{$\mathcal{L}_{\widehat{cr}}$} & \multicolumn{1}{c|}{$\mathcal{L}_{nc}$} & \multicolumn{1}{c|}{$\mathcal{L}_{kcn}$} & $\mathcal{L}_{dcr}$ & \multicolumn{1}{c|}{MAE$\downarrow$}          & \multicolumn{1}{c|}{RMSE$\downarrow$}          & R$\uparrow$          \\ \hline
\multicolumn{1}{c|}{$\checkmark$} &              & \multicolumn{1}{c|}{$\checkmark$}                 & \multicolumn{1}{c|}{}                   & \multicolumn{1}{c|}{}                    &                     & \multicolumn{1}{c|}{12.27}                         & \multicolumn{1}{c|}{13.64}                          & 0.40                     \\ \hline 
\multicolumn{1}{c|}{$\checkmark$} &              & \multicolumn{1}{c|}{$\checkmark$}                 & \multicolumn{1}{c|}{$\checkmark$}       & \multicolumn{1}{c|}{}                    &                     & \multicolumn{1}{c|}{12.33}                         & \multicolumn{1}{c|}{13.64}                          & 0.29                    \\ \hline
\multicolumn{1}{c|}{$\checkmark$} &              & \multicolumn{1}{c|}{$\checkmark$}                 & \multicolumn{1}{c|}{}       & \multicolumn{1}{c|}{$\checkmark$}                    &         & \multicolumn{1}{c|}{12.07}                         & \multicolumn{1}{c|}{13.49}                          & 0.41                      \\ \hline
\multicolumn{1}{c|}{$\checkmark$} &              & \multicolumn{1}{c|}{$\checkmark$}                 & \multicolumn{1}{c|}{}                   & \multicolumn{1}{c|}{}        &     $\checkmark$    & \multicolumn{1}{c|}{10.61}                         & \multicolumn{1}{c|}{11.02}                          & 0.46                    \\ \hline
\multicolumn{1}{c|}{$\checkmark$} &              & \multicolumn{1}{c|}{$\checkmark$}                 & \multicolumn{1}{c|}{$\checkmark$}       & \multicolumn{1}{c|}{$\checkmark$}        & $\checkmark$        & \multicolumn{1}{c|}{8.10}                         & \multicolumn{1}{c|}{9.43}                          & 0.46                     \\ \hline
\multicolumn{1}{c|}{}             & $\checkmark$ & \multicolumn{1}{c|}{$\checkmark$}                 & \multicolumn{1}{c|}{}                   & \multicolumn{1}{c|}{}                    &                     & \multicolumn{1}{c|}{9.79}                         & \multicolumn{1}{c|}{11.66}                          & 0.49                     \\ \hline
\multicolumn{1}{c|}{}             & $\checkmark$ & \multicolumn{1}{c|}{$\checkmark$}                 & \multicolumn{1}{c|}{$\checkmark$}       & \multicolumn{1}{c|}{$\checkmark$}        & $\checkmark$        & \multicolumn{1}{c|}{\textbf{6.01}}                         & \multicolumn{1}{c|}{\textbf{6.79}}                          & \textbf{0.58 }                     \\ \hline
\end{tabular}
\end{table}

\begin{figure} [t]
     \centering
    \begin{tabular}{c} 
     {\includegraphics[width=8cm]{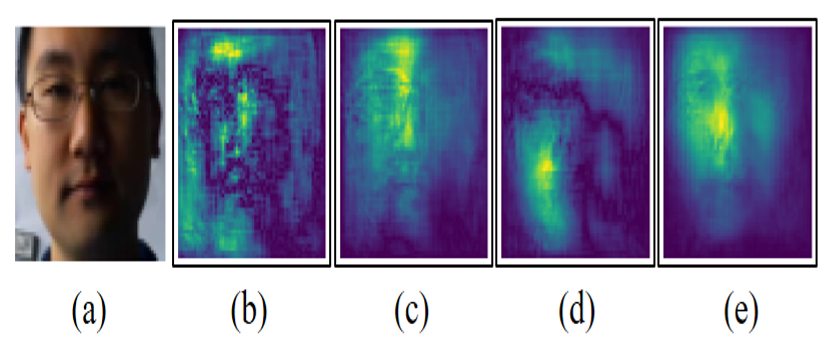}}   
    \end{tabular} 
\caption{  
Visualization of saliency maps \cite{simonyan2013deep} using different combinations of modules and losses: (a) an input facial image, and (b)-(e) the saliency maps generated by the first, fifth, sixth, and seventh cases listed in Table \ref{tab:ablation_different_modules} for cross-testing \textbf{U+P$\rightarrow$C}. Note that, a good rPPG estimator is expected to yield a saliency map with stronger responses in the facial regions \cite{sun2022contrast}.
}
 \label{fig:ablation_visualization} 
\end{figure}

\subsection{Ablation Study}
\subsubsection{
Comparison between Different Loss Terms and Modules
} 
Because \textbf{COHFACE} is a challenging dataset involving significant lighting variations on the facial regions, in Table \ref{tab:ablation_different_modules}, we compare using  different combinations of modules and losses on the protocol of cross-testing  \textbf{{{}}U$+$P{{}}$\rightarrow$C}. 
In addition, in Figure \ref{fig:ablation_visualization}, we visualize their corresponding gradient-based saliency maps \cite{chen2018deepphys,gideon2021way,nowara2021benefit,yu2019remote,sun2022contrast} to compare the  effectiveness of different rPPG models learned by these combinations.
 
First, we compare the performance of interference-carrying rPPG estimation (i.e., 3DCNN + $\mathcal{L}_{\hat{cr}}$) with that of de-interfered rPPG estimation under
various loss combinations, including 3DCNN + $\mathcal{L}_{\hat{cr}}$ + $\mathcal{L}_{nc}$, 3DCNN + $\mathcal{L}_{\hat{cr}}$ + $\mathcal{L}_{kcn}$, 3DCNN + $\mathcal{L}_{\hat{cr}}$ + $\mathcal{L}_{dcr}$, and 3DCNN + $\mathcal{L}_{\hat{cr}} + \mathcal{L}_{nc} + \mathcal{L}_{kcn} + \mathcal{L}_{dcr}$. 
In the second and third cases (i.e., 3DCNN + $\mathcal{L}_{\hat{cr}}$ + $\mathcal{L}_{nc}$ and 3DCNN + $\mathcal{L}_{\hat{cr}}$ + $\mathcal{L}_{kcn}$), the rPPG models, because learning only the characteristics of interferences but without pushing the rPPG signals away from them, show only slight improvement over the first case (3DCNN + $\mathcal{L}_{\hat{cr}}$).
Similarly, in the fourth case (3DCNN + $\mathcal{L}_{\hat{dcr}}$), although the rPPG signals are pushed away from the interferences, the lack of two loss constraints for learning interference characteristics results in little performance improvement compared to the first case (i.e., 3DCNN + $\mathcal{L}_{\hat{cr}}$).
When all losses $\mathcal{L}_{\hat{cr}} + \mathcal{L}_{nc} + \mathcal{L}_{kcn} + \mathcal{L}_{dcr}$ are included, 
the results in Table \ref{tab:ablation_different_modules} show that the fifth case (3DCNN + $\mathcal{L}_{\hat{cr}} + \mathcal{L}_{nc} + \mathcal{L}_{kcn} + \mathcal{L}_{dcr}$), which incorporates de-interfered feature learning to mitigate the interferences within the estimated rPPG signals, significantly improves over the interference-carrying rPPG estimation of the first case. 
Also, as shown in Figures \ref{fig:ablation_visualization} (b) and (c) for the first and fifth cases, respectively, the saliency map of the interference-carrying rPPG estimation (Figure \ref{fig:ablation_visualization} (b)) focuses only on partial facial regions due to challenging illumination variations, whereas the saliency map of the de-interfered rPPG estimation (Figure \ref{fig:ablation_visualization} (c)) more extensively covers even dimly illuminated regions.

Next, we compare the two cases of interference-carrying rPPG estimation using non-descriptive feature learning (i.e., the first case using 3DCNN + $\mathcal{L}_{\hat{cr}}$) with the one using descriptive feature learning (i.e., the sixth case using 3DLDC + $\mathcal{L}_{\hat{cr}}$).
The results in Table \ref{tab:ablation_different_modules} show that, by including descriptive feature learning into DD-rPPGNet, the sixth case substantially outperforms the first one.
Figures \ref{fig:ablation_visualization} (b) and (d) also demonstrate that, because 3DLDC better captures subtle color changes on facial skin under varying reflections, inclusion of descriptive feature learning in Figure \ref{fig:ablation_visualization} (d) indeed yields more accurate saliency maps concentrated on facial regions. 

Finally, when including descriptive feature learning into de-interfered rPPG estimation in DD-rPPGNet (i.e., the seventh case using 3DLDC + $\mathcal{L}_{\hat{cr}} + \mathcal{L}_{nc}+\mathcal{L}_{kcn}+\mathcal{L}_{dcr} $), we see that the performance in the last row of Table \ref{tab:ablation_different_modules} and the saliency maps in Figure \ref{fig:ablation_visualization} (e) are greatly improved over the other cases and demonstrate the effectiveness of the proposed DD-rPPGNet.

\begin{table}[t]
\caption{Ablation study on the protocol of cross-testing  \textbf{{{}}U$+$P{{}}$\rightarrow$C}, using different convolution kernels.}
\label{tab:ablation_different_convolution}
\setlength{\tabcolsep}{2.5pt} 
\footnotesize 
\centering
\begin{tabular}{c|ccc|c|c|c}
\hline
\multirow{2}{*}{Method}                      & \multicolumn{3}{c|}{$\textbf{U}+\textbf{P}\rightarrow\textbf{C}$}                          & \multirow{2}{*}{$\#param.$} & \multirow{2}{*}{FLOPs} & \multirow{2}{*}{FPS} \\ \cline{2-4}
                                             & \multicolumn{1}{c|}{MAE$\downarrow$} & \multicolumn{1}{c|}{RMSE$\downarrow$} & R$\uparrow$ &                             &                        &                      \\ \hline
Vanilla 3DCNN                                & \multicolumn{1}{c|}{8.10}            & \multicolumn{1}{c|}{9.43}             & 0.46        &               1.33              &       52.25                 &            689.59          \\ \hline
TDC \cite{yu2020autohr}     & \multicolumn{1}{c|}{6.64}            & \multicolumn{1}{c|}{8.27}             & 0.47        &      1.33                       &           52.36             &     678.01                 \\ \hline
3DCDC  \cite{zhao2021video} & \multicolumn{1}{c|}{6.84}            & \multicolumn{1}{c|}{8.25}             & 0.52        &        1.33            &           52.36             &         685.47             \\ \hline
3DLDC                                        & \multicolumn{1}{c|}{\textbf{6.01}}            & \multicolumn{1}{c|}{\textbf{6.79}}             & \textbf{0.58}        &        2.26                     &         52.25               &      686.75                \\ \hline
\end{tabular}
\end{table}

\subsubsection{Comparison between Different Convolutions for Learning Descriptive Features }   
In Table \ref{tab:ablation_different_convolution}, we compare using different 3D convolutions, including  vanilla 3DCNN, TDC \cite{yu2020autohr}, and 3DCDC \cite{zhao2021video}, to replace the proposed 3DLDC in DD-rPPGNet under the same training loss $\mathcal{L}_{\hat{cr}} + \mathcal{L}_{nc}+\mathcal{L}_{kcn}+\mathcal{L}_{dcr} $ and then test on the protocol \textbf{{{}}U$+$P{{}}$\rightarrow$C}. 
The results in Table \ref{tab:ablation_different_convolution} show that 3DLDC substantially outperforms the others and demonstrates that the proposed learnable descriptor successfully captures subtle temporal and color changes for learning descriptive rPPG features. 
In addition, in Table \ref{tab:ablation_different_convolution}, we compare the model size, computational complexity, and inference throughput of DD-rPPGNet using various metrics, including the number of parameters (\#param.), FLOPs, and FPS. The results indicate that, although the proposed 3DLDC incorporates additional learnable descriptors, primarily consisting of vanilla and enhanced convolutions similar to TDC \cite{yu2020autohr} and 3DCDC \cite{zhao2021video}, it maintains a relatively fast inference speed compared to other methods.

\begin{table}[t] 
\caption{
{Ablation study on different interference conditions.}
}
\label{tab:inject_noise}
\setlength{\tabcolsep}{6pt} 
\scriptsize 
\centering 
\begin{tabular}{ >{\centering\arraybackslash}l c c c c c c }
    \toprule 
    \multirow{2}{*}{Methods} 
    & \multirow{2}{*}{\shortstack{P.}}
    & \multirow{2}{*}{\shortstack{Interference\\Type}}
    & \mytab{MAE$\downarrow$\\(bpm)} 
    & \mytab{RMSE$\downarrow$\\(bpm)}
    & \multirow{2}{*}{\mytab{R$\uparrow$}}
    \\
    \addlinespace[1pt]
    \hline
    \addlinespace[2pt] 
    Gideon2021 \cite{gideon2021way} & \multirow{4}{*}{1} & \multirow{4}{*}{\shortstack{Head\\motion}} & 1.89 & 2.81 & 0.986 \\
    Contrast-Phys \cite{sun2022contrast} & & & 1.59 & 2.69 & 0.991 \\ 
    SiNC \cite{speth2023non} & & & 0.67 & 1.38 & 0.996\\
    \textbf{DD-rPPGNet} & & & \textbf{0.51} & \textbf{1.09} & \textbf{0.998} \\
    \addlinespace[1pt]
    \hline
    \addlinespace[2pt]
    Gideon2021 \cite{gideon2021way} & \multirow{4}{*}{2} & \multirow{4}{*}{\shortstack{Facial\\expression}} & 1.85 & 2.93 & 0.969\\
    Contrast-Phys \cite{sun2022contrast} & & & 1.60 & 2.68 & 0.978 \\ 
    SiNC \cite{speth2023non} & & & 1.32 & 4.45 & 0.956\\
    \textbf{DD-rPPGNet } & & & \textbf{0.25} & \textbf{0.43} & \textbf{0.983} \\
    \addlinespace[1pt]
    \hline
    \addlinespace[2pt]
    Gideon2021 \cite{gideon2021way} & \multirow{4}{*}{3} & \multirow{4}{*}{\shortstack{Compression\\ artifact}} & 11.14 & 15.68 & 0.216\\
    Contrast-Phys \cite{sun2022contrast} & & & 8.82 & 10.29 & 0.357 \\ 
    SiNC \cite{speth2023non} & & & 14.51 & 17.27 & 0.043\\
     \textbf{DD-rPPGNet } & & & \textbf{3.41} & \textbf{5.10} & \textbf{0.740} \\
    \addlinespace[1pt]
    \hline
    \addlinespace[2pt] 
    Gideon2021 \cite{gideon2021way} & \multirow{4}{*}{4} & \multirow{4}{*}{\makebox[1.3cm]{\shortstack{ Illumination \\ variations }}} & 13.79 & 16.08 & 0.177\\
    Contrast-Phys \cite{sun2022contrast} & & & 23.02 & 27.23 & 0.104 \\ 
    SiNC \cite{speth2023non} & & & 18.86 & 23.11 & 0.050\\
    \textbf{DD-rPPGNet } & & & \textbf{8.54} & \textbf{8.86} & \textbf{0.464} \\
    \addlinespace[1pt]
    \hline
    \addlinespace[2pt]
    Gideon2021 \cite{gideon2021way} & \multirow{4}{*}{5} & \multirow{4}{*}{\makebox[1.3cm]{\shortstack{Periodic noise\\ injected in\\ non-facial and \\facial regions}}} & 21.78 & 24.89 & 0.142\\
    Contrast-Phys \cite{sun2022contrast} & & & 7.92 & 13.62 & 0.707 \\ 
    SiNC \cite{speth2023non} & & & 18.69 & 23.39 & 0.106\\
    \textbf{DD-rPPGNet } & & & \textbf{0.92} & \textbf{1.70} & \textbf{0.994} \\
    \bottomrule
\end{tabular}
 \label{tab:ablation_various_noise} 
\end{table}

\subsubsection{Effectiveness of De-interfered rPPG Estimation}

To evaluate the effectiveness of the proposed de-interfered rPPG estimation, we design five protocols of different interference conditions, including head motion, facial expressions, compression artifacts, lighting variations, and periodic noises, and conduct experimental comparisons in Table \ref{tab:ablation_various_noise}. Detail description of these protocols is given in the supplementary material.
As shown in Table \ref{tab:ablation_various_noise}, in the cases of protocols P1 and P2, the proposed DD-rPPGNet remains resilient to the interference caused by head motion and expression changes and outperforms previous unsupervised rPPG estimation methods \cite{gideon2021way,sun2022contrast,speth2023non}. Furthermore, in the cases of challenging interference such as compression artifacts, illumination variations, and periodic noises in P3-P5,
we observe that these methods \cite{gideon2021way,sun2022contrast,speth2023non} have a significant decrease in performance; whereas DD-rPPGNet successfully resists the interference, thanks to the proposed de-interfered rPPG estimation, and largely outperforms these methods.


\begin{table*}[htb]
\centering 
\setlength\tabcolsep{0pt}
\renewcommand{\arraystretch}{1.1} 
\caption{
Comparison of intra-domain testing on \textbf{U},  \textbf{P}, \textbf{C(Natural)}, \textbf{Car}, and \textbf{V}.}
\label{tab:intra_testing}
\scalebox{0.8}{
\small
\begin{tabular*}{\textwidth}{@{\extracolsep{\fill}}llcccccccccccccc{c}}
    \toprule
    \multirow{3}{*}{\parbox{1cm}{Method\\Types}} 
    & \multirow{3}{*}{\mytab{Methods}} 
    & \mc{3}{\mytab{\textbf{U}}} 
    & \mc{3}{\mytab{\textbf{P}}}
    & \mc{3}{\mytab{\textbf{C(Natural)}}} 
    & \mc{3}{\mytab{\textbf{Car(RGB)}}} 
    & \mc{3}{\mytab{\textbf{V}}} \\
    \cmidrule{3-5} \cmidrule{6-8} \cmidrule{9-11}
    \cmidrule{12-14}  \cmidrule{15-17} 
    &
    & \mytab{MAE$\downarrow$\\(bpm)} 
    & \mytab{RMSE$\downarrow$\\(bpm)}
    & \multirow{2}{*}{\mytab{R$\uparrow$}}
    & \mytab{MAE$\downarrow$\\(bpm)}
    & \mytab{RMSE$\downarrow$\\(bpm)}
    & \multirow{2}{*}{\mytab{R$\uparrow$}}
    & \mytab{MAE$\downarrow$\\(bpm)}
    & \mytab{RMSE$\downarrow$\\(bpm)}
    & \multirow{2}{*}{\mytab{R$\uparrow$}}
    & \mytab{MAE$\downarrow$\\(bpm)}
    & \mytab{RMSE$\downarrow$\\(bpm)}
    & \multirow{2}{*}{\mytab{R$\uparrow$}}
    & \mytab{MAE$\downarrow$\\(bpm)}
    & \mytab{RMSE$\downarrow$\\(bpm)}
    & \multirow{2}{*}{\mytab{R$\uparrow$}}\\
    \midrule
    \multirow{7}{*}{\parbox{1cm}{Super-\\vised}} 
    & DeepPhys \cite{chen2018deepphys} & - & - & - & - & - & - & - & - & - & - & - & - & 11.0 & 13.8 & 0.11 \\
    & AutoHR \cite{yu2020autohr} & - & - & - & - & - & - & - & - & - & - & - & - & 5.68 & 8.68 & 0.72  \\
    & PulseGAN \cite{song2021pulsegan} & 1.19 & 2.10 & \underline{0.98} & - & - & - & - & - & - & - & - & - & - & - & -  \\
    & Dual-GAN \cite{lu2021dual} & 0.44 & 0.67 & \textbf{0.99} & 0.82 & 1.31 & \underline{0.99} & - & - & - & - & - & - & \textbf{4.93} & \textbf{7.68} & \textbf{0.81} \\
    & RErPPG-Net \cite{hsieh2022augmentation} & 0.41 & 0.56 & \textbf{0.99} & 0.38 & 0.54 & 0.96 & {11.56} & {12.41} & 0.13 & 18.84 & 21.63 & 0.10 & - & - & - \\
    & PhysFormer \cite{yu2022physformer} & \underline{0.33} & \underline{0.41} & \textbf{0.99} & \underline{0.11} & \underline{0.20} & \underline{0.99} & \underline{10.36} & \underline{11.71} & 0.14 & \underline{15.47} & \underline{17.54} & 0.09 & \underline{4.97} & \underline{7.79} & \underline{0.78} \\
    \addlinespace[1pt]
    \hline
    \addlinespace[2pt]
    \multirow{5}{*}{\shortstack{Unsup-\\ervised}} 
    & Gideon2021 \cite{gideon2021way} & 1.85 & 4.28 & 0.93 & 2.3 & 2.9 & \underline{0.99} & {13.79} & {16.08} & \underline{0.18} & 17.76 & 18.77 & 0.11 & 25.84 & 28.20 & 0.47 \\
    & Contrast-Phys \cite{sun2022contrast} & 0.64 & 1.00 & \textbf{0.99} & 1.00 & 1.40 & \underline{0.99} & 23.02 & 27.23 & 0.10 & 19.32 & 21.30 & 0.09 & 25.13 & 30.04 & 0.48\\
    & SiNC \cite{speth2023non} & 0.59 & 1.83 & \textbf{0.99} & 0.61 & 1.84 & \textbf{1.00} & 18.86 & 23.11 & 0.05 & 15.83 & 20.56 & \underline{0.14} & 14.33 & 22.83 & 0.45\\
    & Yue \textit{et al}. \cite{yue2023facial} & 0.58 & 0.94 & \textbf{0.99} & 1.23 & 2.01 & \underline{0.99} & - & - & - & - & - & - & - & - & -\\
    & \textbf{DD-rPPGNet} & \textbf{0.19} & \textbf{0.22} & \textbf{0.99} & \textbf{0.10} & \textbf{0.18} & \textbf{1.00} & \textbf{8.54} & \textbf{8.86} & \textbf{0.46} & \textbf{13.93} & \textbf{15.14} & \textbf{0.18} & 13.53 & 18.85 & 0.56 \\
    
    \bottomrule
\end{tabular*}}
\end{table*}


\begin{table*}[htb]
\centering
\setlength\tabcolsep{0pt}
\renewcommand{\arraystretch}{1.0} 
\caption{Comparison of cross-domain testing on  \mytab{\textbf{P}$+$\textbf{C }$\rightarrow$\textbf{U}}, \mytab{\textbf{U}$+$\textbf{C }$\rightarrow$\textbf{P}}, and 
\mytab{\textbf{U}$+$\textbf{P}$\rightarrow$\textbf{C }}.
}
\label{tab:cross_testing}
\scalebox{0.8}{
\centering
\begin{tabular*}{\textwidth}{@{\extracolsep{\fill}}llcccccccc{c}}
    \toprule
    \multirow{3}{*}{\parbox{1cm}{Method\\Types}} 
    & \multirow{3}{*}{\mytab{Methods}} 
    & \mc{3}{\mytab{\textbf{P}$+$\textbf{C }$\rightarrow$\textbf{U}}}
    & \mc{3}{\mytab{\textbf{U}$+$\textbf{C }$\rightarrow$\textbf{P}}}
    & \mc{3}{\mytab{\textbf{U}$+$\textbf{P}$\rightarrow$\textbf{C }}}  \\
    \cmidrule{3-5} \cmidrule{6-8} \cmidrule{9-11}
    &
    & \mytab{MAE$\downarrow$\\(bpm)}
    & \mytab{RMSE$\downarrow$\\(bpm)}
    & \multirow{2}{*}{\mytab{R$\uparrow$}}
    & \mytab{MAE$\downarrow$\\(bpm)}
    & \mytab{RMSE$\downarrow$\\(bpm)}
    & \multirow{2}{*}{\mytab{R$\uparrow$}}
    & \mytab{MAE$\downarrow$\\(bpm)}
    & \mytab{RMSE$\downarrow$\\(bpm)}
    & \multirow{2}{*}{\mytab{R$\uparrow$}}\\
    \midrule
    \multirow{5}{*}{\parbox{1cm}{Super-\\vised}} 
    & Multi-task \cite{tsou2020multi} & 1.06 & 2.70 & - & 4.24 & 6.44 & - & - & - & - \\
    & Dual-GAN \cite{lu2021dual} & 0.74 & \underline{1.02} & \textbf{0.99} & - & - & - & - & - & -  \\
    & DG-rPPGNet \cite{chung2022domain} & \underline{0.63} & 1.35 & 0.88 & 3.02 & 4.69 & 0.88 & \underline{7.19} & 8.99 & 0.30 \\
    & RErPPG-Net \cite{hsieh2022augmentation} & \textbf{0.12} & \textbf{0.17} & \textbf{0.99} & \underline{0.17} & \underline{0.20} & \textbf{0.99} & 7.26 & 9.27 & 0.31  \\
    & PhysFormer \cite{yu2022physformer} & 0.85 & 1.48 & 0.83 & 0.23 & 0.40 & 0.91 & 7.29 & \underline{8.60} & \underline{0.48}  \\
    \addlinespace[1pt]
    \hline
    \addlinespace[2pt]
    \multirow{4}{*}{\shortstack{Unsup-\\ervised}} 
    & Gideon2021 \cite{gideon2021way}  & 7.69 & 8.86 & 0.97 & 0.83 & 1.68 & \underline{0.98} & 10.14 & 11.39 & 0.34  \\ 
    & Contrast-Phys \cite{sun2022contrast} & 1.30 & 2.01 & \underline{0.98} & 0.36 & 0.73 & \textbf{0.99} & 8.73 & 11.93 & 0.43 \\ 
    & SiNC \cite{speth2023non} & 17.00 & 22.01 & 0.33 & 30.33 & 34.17 & -0.25 & 18.56 & 25.83 & 0.13\\
    & \textbf{DD-rPPGNet} & 0.90 & 1.27 & \textbf{0.99} & \textbf{0.08} & \textbf{0.11} & \textbf{0.99} & \textbf{6.01} & \textbf{6.79} & \textbf{0.58} \\
    \bottomrule\\ 

\end{tabular*}}
\end{table*}

\subsection{Intra-Domain and Cross-Domain Testing}
 
In Tables \ref{tab:intra_testing} and \ref{tab:cross_testing}, we conduct intra-domain and cross-domain testing  and compare the results with previous supervised and unsupervised methods. 

\subsubsection{Intra-Domain Testing}  
\begin{figure} [t]
 \centering
   \begin{minipage}[b]{1\linewidth} 
    \includegraphics[width=8.5cm]{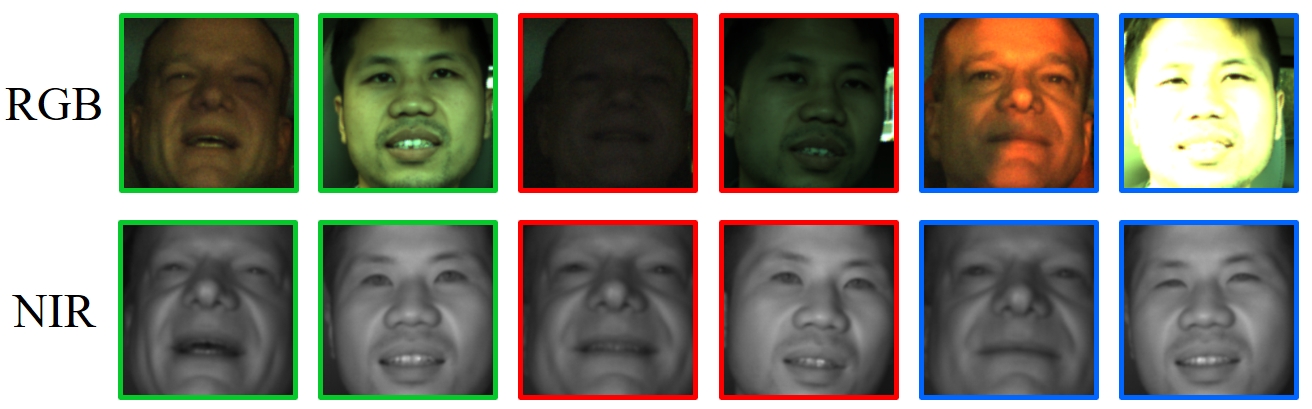} 
  \end{minipage}
\caption{  
Sample images from \textbf{MR-NIRP Car} \cite{NowaraDriving} captured under  different illumination conditions, including normal illumination (boxes in green), dark  illumination (boxes in red), and bright illumination (boxes in blue).} 
 \label{fig:driving_dataset}   
\end{figure}

In Table \ref{tab:intra_testing}, we conduct intra-domain testing on five rPPG datasets, \textbf{U}, \textbf{P}, \textbf{C(Natural)}, \textbf{Car}, and \textbf{V}, where \textbf{U} and \textbf{P} involve minor interference and the other three datasets \textbf{C(Natural)}, \textbf{Car}, and \textbf{V} are more challenging with significant illumination variations.  
First, we see that DD-rPPGNet outperforms all the previous supervised and unsupervised rPPG estimation methods on \textbf{U} and \textbf{P}  under minor interference.
Next, when facing challenging datasets \textbf{C(Natural)}, \textbf{Car}, and \textbf{V}, DD-rPPGNet still outperforms all the previous unsupervised methods and achieves comparable performances to the state-of-the-art supervised method.
In particular, as shown in Figure \ref{fig:ablation_visualization} (e), DD-rPPGNet effectively tackles the challenges posed by lighting variations in \textbf{C(Natural)} through de-interfered rPPG estimation and descriptive feature learning.
For the large-scale dataset \textbf{V}, which is collected across nine distinct scenarios and encompasses various variations in face pose, scale, and illumination conditions, previous unsupervised methods have a noticeable performance drop, whereas the proposed DD-rPPGNet remains and continues to yield promising results. 
As to the RGB videos in the driving dataset \textbf{Car} \cite{NowaraDriving}, the major challenge stems from extreme illumination variations, ranging from very dark to very bright conditions, as shown in Figure~\ref{fig:driving_dataset}. Under these conditions, interference signals extracted from background clips and rPPG signals extracted from foreground clips exhibit strong positive correlations, as illustrated in Figure~\ref{fig:local similarity in dataset}. Consequently, the \textbf{Car} dataset contains significant interference and poses greater challenges than the others.
Nevertheless, DD-rPPGNet continues to deliver promising performance and robustness, even in the presence of substantial interferences and challenging conditions.

\subsubsection{Cross-Domain Testing} 
In Table \ref{tab:cross_testing}, we follow the benchmark proposed in \cite{chung2022domain} to conduct cross-domain testing on \textbf{U}, \textbf{P}, and \textbf{C} by using two datasets for training and the remaining one for testing. 
The results in Table \ref{tab:cross_testing} show that DD-rPPGNet yields the best results on the protocols \mytab{\textbf{U}$+$\textbf{C}$\rightarrow$\textbf{P}} and \mytab{\textbf{U}$+$\textbf{P}$\rightarrow$\textbf{C}}, and achieves comparable performance with the state-of-the-art method RErPPG-Net \cite{hsieh2022augmentation} on \mytab{\textbf{P}$+$\textbf{C}$\rightarrow$\textbf{U}}.
Note that, RErPPG-Net \cite{hsieh2022augmentation}, by using ground truth rPPG signals to augment the training data, indeed performs well when there exists little domain gap between the training and testing domains, such as the protocols \mytab{\textbf{P}$+$\textbf{C}$\rightarrow$\textbf{U}} and \mytab{\textbf{U}$+$\textbf{C}$\rightarrow$\textbf{P}}. However, RErPPG-Net \cite{hsieh2022augmentation} performs relatively poorly on the protocol \mytab{\textbf{U}$+$\textbf{P}$\rightarrow$\textbf{C}} because of the significant difference between the training and testing domains.
By contrast, with the de-interfered rPPG estimation and descriptive feature learning, the proposed DD-rPPGNet successfully achieves stable and promising results even on the challenging cross-domain protocol \mytab{\textbf{U}$+$\textbf{P}$\rightarrow$\textbf{C(Natural)}}.

\begin{figure} [t]
 \centering
   \begin{minipage}[b]{0.7\linewidth} 
    \includegraphics[width=5.cm]{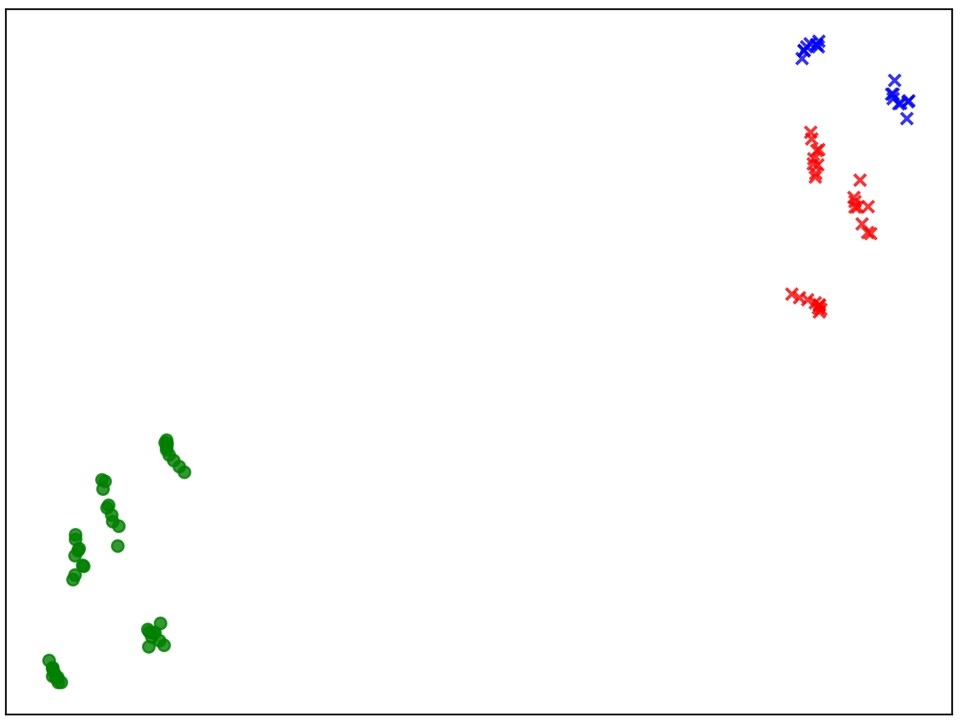} 
  \end{minipage}
\caption{  
 $t$-SNE visualizations on \textbf{PURE} \cite{stricker2014non} for rPPG features and different interference features.  (green dots: rPPG features, red crosses: interference features of head motion, and blue crosses: interference features of compression artifact).
 } 
 \label{fig:different features}   
\end{figure} 

\subsection{$t$-SNE Visualization}
\subsubsection{ 
Comparison Between Using Only Original Facial Videos and Augmented Facial Videos} 
 
In Figure~\ref{fig:different rPPG features}, we extract rPPG features from both the original facial video and its augmented version using the off-the-shelf rPPG estimator \cite{sun2022contrast} and show the $t$-SNE visualization for different HR ranges. The results show that rPPG features corresponding to similar HRs, whether from the original facial video or its augmented version, tend to cluster together. This observation aligns with the conclusion in \cite{yang2023simper} that rPPG features associated with similar HRs tend to form compact clusters in the latent feature space.

\subsubsection{ 
Comparison between rPPG Features and Interference Features}
 
In Figure~\ref{fig:different features}, we show the $t$-SNE visualization of rPPG and interference features using the trained DD-rPPGNet on Protocols 1 and 3 from Table~\ref{tab:ablation_various_noise}, where the videos are affected by head motion and compression artifacts. The results demonstrate that the proposed DD-rPPGNet effectively clusters interference features of the same type together, while keeping rPPG features distinctly separated from these interference features. 

\subsection{ Limitation and Future Work }
 
The performance of rPPG estimation from facial videos remain limited by challenging illumination conditions. Since RGB images are highly sensitive to extreme lighting—whether too dark or too bright, our method, like existing rPPG estimation methods
 that rely solely on the RGB modality, also suffers from unsatisfactory results under these conditions.
For example, as shown in Figure~\ref{fig:driving_dataset}, the driving rPPG dataset \textbf{MR-NIRP Car} \cite{NowaraDriving} includes both RGB and Near-Infrared (NIR) images captured under diverse and challenging illumination conditions.
The intra-testing results in Table \ref{tab:intra_testing} demonstrate that using only RGB images is insufficient for accurate heart rate estimation under these conditions. Therefore, further exploration of incorporating multiple modalities \cite{liu2023information,park2022self} should be crucial to address the challenges posed by extreme illumination in rPPG estimation.  
In particular, since different modalities exhibit distinct characteristics, integrating the proposed 3DLDC into each modality separately is expected to better capture modality-specific temporal features compared to using a shared 3DLDC across different modalities.
In addition, further investigation into the dynamic adaptation of 3DLDC across various scenarios and modalities is worth pursuing to enhance the overall robustness of multi-modal rPPG estimation in complex environments.

\section{Conclusion} 

In this paper, we propose a novel fully unsupervised rPPG estimation network, called DD-rPPGNet, to tackle the challenges of rPPG estimation in the presence of interference.
First, we investigate the local spatial-temporal characteristics of interference and propose an interference estimation method to model the interference in rPPG signals.
Next, we propose a de-interfered rPPG estimation method by enforcing the rPPG signals to be distinct from the estimated interference.
Furthermore, we propose an effective descriptive rPPG feature learning to capture the subtle color changes on skin to enhance rPPG estimation.  
 Extensive experiments demonstrate that DD-rPPGNet outperforms previous unsupervised rPPG estimation methods and achieves competitive performance with state-of-the-art supervised rPPG estimation method.

\section*{Acknowledgment}  
This work was supported in part by the National Science and Technology Council  grants 111-2221-E-007-109-MY3 and 112-2221-E-007-082-MY3 of Taiwan, and the National Natural Science Foundation of China under Grant 62272103. 
The authors would like to express their grateful appreciation to the associate editor and the anonymous reviewers for their valuable efforts in improving the quality of this article.

\bibliographystyle{IEEEtran}
\bibliography{egbib} 

 
\begin{IEEEbiography} 
[{\includegraphics[width=1in,height=1.25in,clip,keepaspectratio]{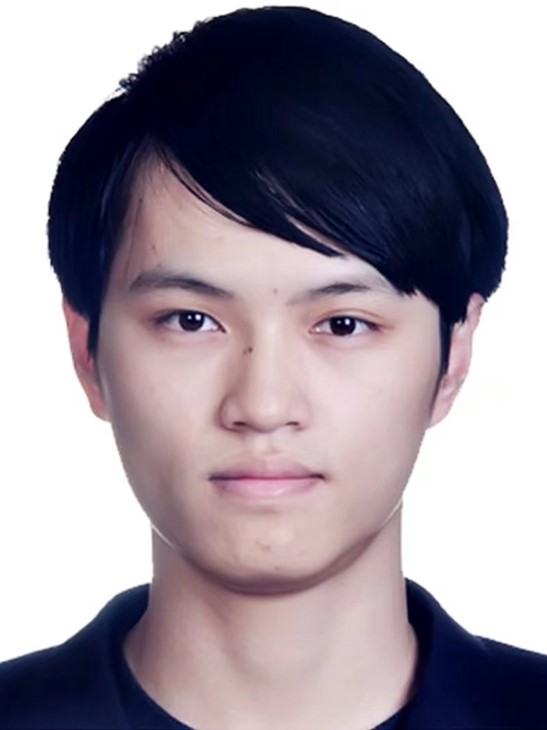}}]{Pei-Kai Huang} 
received the Ph.D. degree in Computer Science from National Tsing Hua University in 2025, the M.S. degree in Computer Science from National Central University in 2019, and the B.S. degree in Software Engineering from Fujian Normal University, Fujian, China, in 2017.
His research interests include multimedia content security, biometrics, and computer vision. 
As the first author, he has had 12 technical papers published at prestigious conferences, including CVPR, AAAI, BMVC, ICME, and ICIP, as well as in renowned journals, such as TIFS and Information Sciences. 
\end{IEEEbiography} 
 
\begin{IEEEbiography} 
[{\includegraphics[width=1in,height=1.25in,clip,keepaspectratio]{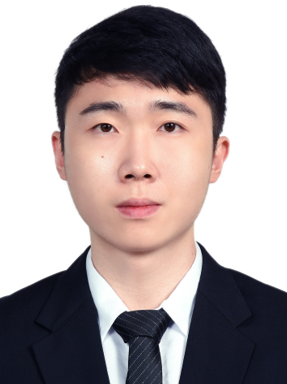}}]{Tzu-Hsien Chen} 
received the B.S. degree in Computer Science from National Chengchi University, Taiwan, in 2022,  
and the M.S. degree in Computer Science from National Tsing Hua University, Taiwan, in 2024.  
His research interets include computer vision, deep learning, remote photoplethysmography (rPPG) estimation, and face anti-spoofing.
\end{IEEEbiography}

\begin{IEEEbiography} 
[{\includegraphics[width=1in,height=1.25in,clip,keepaspectratio]{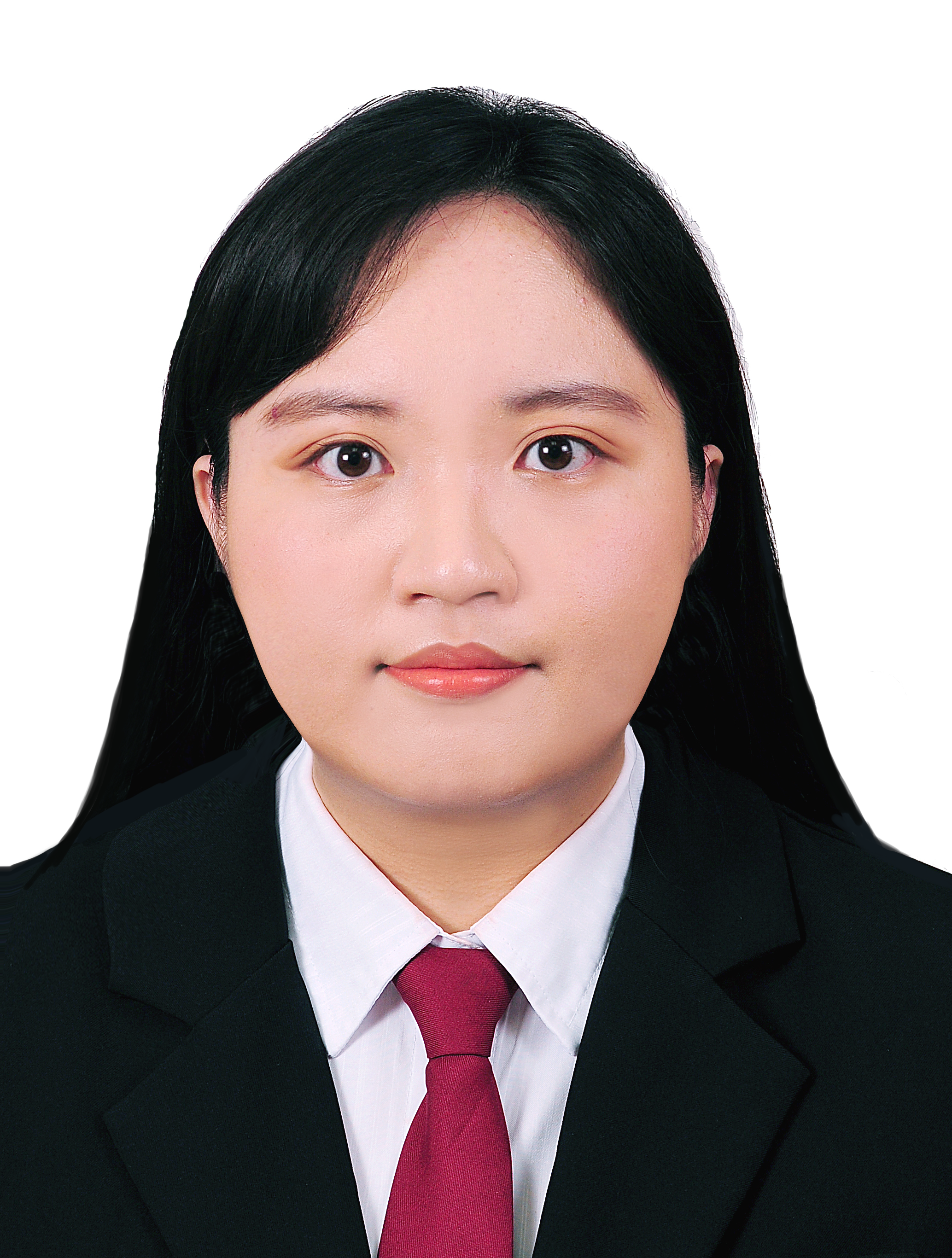}}]{Ya-Ting Chan} 
received the B.S. degree in Com-
puter Science from Yuan Ze University, Taiwan, in 2023. She is currently pursuing the M.S. degree in Computer Science at National Tsing Hua University, Taiwan. Her research interests include computer vision, deep learning, multi-modal learning, physiological signal estimation, and face anti-spoofing.
\end{IEEEbiography}

\begin{IEEEbiography} 
[{\includegraphics[width=1in,height=1.25in,clip,keepaspectratio]{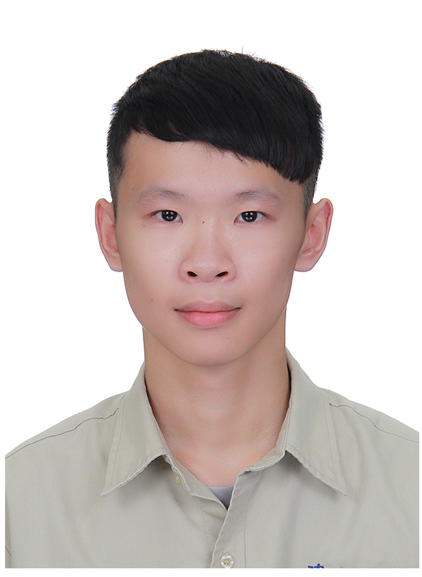}}]{Kuan-Wen Chen} 
received the B.S. degree in Computer Science and Information Engineering from National Taiwan Normal University, Taiwan, in 2023. He is currently pursuing the M.S. degree in Computer Science at National Tsing Hua University, Taiwan. His research interests include computer vision, deep learning, and remote photoplethysmography (rPPG) estimation.
\end{IEEEbiography}

\begin{IEEEbiography} 
[{\includegraphics[width=1in,height=1.25in,clip,keepaspectratio]{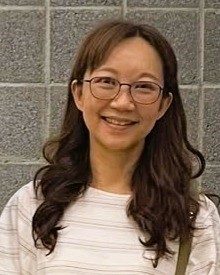}}]{Chiou-Ting Hsu} (M’98–SM’13) received the Ph.D. degree in computer science and information engineering from National Taiwan University (NTU), Taipei, Taiwan, in 1997. From 1998 to 1999, she was with Philips Innovation Center, Taipei, Philips Research, as a Senior Research Engineer. She joined the Department of Computer Science, National Tsing Hua University, Hsinchu, Taiwan, as an Assistant Professor, in 1999, and is currently a Professor. Her research interests include image processing, image and video analysis, pattern recognition, and machine learning. She has served in the editorial board of Advances in Multimedia (2006-2012), IEEE Transactions on Information Forensics and Security (2012-2015), Journal of Visual Communication and Image Representation (2015-2025), and EURASIP Journal on Image and Video Processing (2015-2025). She is currently serving as  IEEE SPS Regional Director-at-large (2024-2025) for Region 10.
\end{IEEEbiography}

\end{document}